\documentclass{elsarticle}

\usepackage{lineno,hyperref}
\usepackage{lineno,hyperref}
\usepackage[T1]{fontenc}
\usepackage{amsfonts}
\usepackage{graphicx,times,amsmath}
\usepackage{subfigure}
\usepackage{algorithm}
\usepackage{algorithmic}
\usepackage{subfigure}
\usepackage[utf8]{inputenc}
\usepackage{textcomp}
\usepackage{eurosym}
\usepackage{color}
\usepackage{url}

\newtheorem{theorem}{Theorem}[section]
\newtheorem{definition}[theorem]{Definition}

\journal{Information Fusion}

\begin{document}

\begin{frontmatter}

\title{A Heuristically Self-Organised Linguistic Attribute Deep Learning in Edge Computing For IoT Intelligence}
%\tnotetext[titlenote]{}

\author{Hongmei He\fnref{footnote1}}
\address{School of Aerospace, Transport and Manufacturing, Cranfield University, Cranfield, UK}
\fntext[footnote1]{Dr Hongmei He is with Cranfield University, Uk, email:h.he@cranfield.ac.uk or maryhhe@gmail.com}

\author{Zhenhuan Zhu\fnref{footnote2}}
\address{Smart Perception Lab, Milton Keynes, UK}
\fntext[footnote2]{Dr Zhenhuan Zhu is with Smart Perception Lab, UK, email:spl.z.zhu@outlook.com or zhenhuan2001@hotmail.com}

%%% Group authors per affiliation:
%\author{Erkki Makinen\fnref{footnote3}}
%\address{School of Information Sciences, University of Tampere, Tampere, Finland}
%\fntext[footnote3]{Prof Erkki M$\ddot{a}$kinen is a Professor of Computer Science at University of Tampere, Finland, email: erkki.makinen@uta.fi}

\begin{abstract}
With the development of Internet of Things (IoT), IoT intelligence becomes emerging technology. “Curse of Dimensionality” is the barrier of data fusion in edge devices for the success of IoT intelligence. Deep learning has attracted great attention recently, due to the successful applications in several areas, such as image processing and natural language processing. However, the success of deep learning benefits from GPU computing. A Linguistic Attribute Hierarchy (LAH), embedded with Linguistic Decision Trees (LDTs) can represent a new attribute deep learning. In contrast to the conventional deep learning, an LAH could overcome the shortcoming of missing interpretation by providing transparent information propagation through the rules, produced by LDTs in the LAH. Similar to the conventional deep learning, the computing complexity of optimising LAHs blocks the applications of LAHs.

In this paper, we propose a heuristic approach to constructing an LAH, embedded with LDTs for decision making or classification by utilising the distance correlations between attributes and between attributes and the goal variable.  The set of attributes is divided to some attribute clusters, and then they are heuristically organised to form a linguistic attribute hierarchy. The proposed approach was validated with some benchmark decision making or classification problems from the UCI machine learning repository. The experimental results show that the proposed self-organisation algorithm can construct an effective and efficient linguistic attribute hierarchy. Such a self-organised linguistic attribute hierarchy embedded with LDTs can not only efficiently tackle ‘curse of dimensionality’ in a single LDT for data fusion with massive attributes, but also achieve better or comparable performance on decision making or classification, compared to the single LDT for the problem to be solved. The self-organisation algorithm is much efficient than the Genetic Algorithm in Wrapper for the optimisation of LAHs. This makes it feasible to embed the self-organisation algorithm in edge devices for IoT intelligence.
\end{abstract}


\end{frontmatter}

%\linenumbers

\section{Introduction}\label{sec:Intro}
The Internet-of-Things (IoT) provides us with a large amount
of sensor data. However, the data by themselves do not provide
value unless we can turn them into actionable and/or contextualized information. Big data analysis allows us to gain new insights by batch-processing and off-line analysis. Currently, a microprocessor-based sensor node can support many channels
(e.g. a Microchip processor can support up to 49 channel
inputs \cite{datasheet2017}). Real-time sensor data analysis and decision-making is preferably automated on-board of IoT devices, which will make IoT intelligence towards reality.

Although the computing capability of a microprocessor has
improved very much, the ‘curse of dimensionality’ is still
a big challenge in data driven machine intelligence, as the
computing complexity of designed model function increases as
the increasing of input space. Blum and Rivest \cite{Blum1992} have proved
that training a 2-layer, 3-nodes and n inputs neural network
is NP-Complete. Obviously, the big barrier of blocking the
applications of deep-learning is the computing complexity,
although it shows great attractive on solving complex nonlinear
problems. With the strong capability of GPU, deep
learning for 2-20 depth networks is successful (e.g. Google
AlphaGo). To save the cost, an edge device of IoT systems
may not need to equip with GPUs, if the machine intelligence
algorithm inside the device is efficient enough. Hence, the
performance improvement of computational intelligence is
continuous work, especially for creating effective and efficient
computing model in edge devices to implement IoT intelligence.

Linguistic decision tree (LDT), a probabilistic tree, has been
well used for decision making and classification problems.
Given an input space of n attributes, each of which can be
described with limit labels. An LDT consists of a set of
attribute nodes and a set of edges. An edge, linking from an
attribute node, represents a label expression that describes the
attribute node (see Section \ref{sec:LDT-LS}). However, the branch number
of a decision tree exponentially increases as the number of
input attributes increases. This shortcoming of an LDT greatly
blocks its applications.

A hierarchical model could help overcome the ``Curse of Dimensionality'' in fuzzy rule-based learning \cite{Raju1991}. Campello and Amaral \cite{Campello2006} provided a cascade hierarchy of  sub-models, using fuzzy relational equations \cite{Pedrycs1993}, and  unilaterally transformed the cascade models into the mathematically equivalent non-hierarchical one. Lawry and He\cite{Lawry2008} proposed a linguistic attribute hierarchy (LAH).  It is a hierarchy of LDTs, each of which has a reduced input space, and represents different functions in the problem space.  However, as the relationship between inputs and the output in the whole problem space could be strong non-linear and uncertain, different LAHs will have different performance for the problem to be solved.  All of the research in \cite{Campello2006,Lawry2008} neither investigated the performance of the proposed hierarchies, nor studied how the hierarchies can be constructed optimally.

He and Lawry \cite{He2009a,He2009b} investigated cascade LAHs, and developed a genetic wrapper algorithm (GWA) to optimise the cascade LAHs.  It was shown that a cascade hierarchy can derive much less rules, compared to the single decision tree. However, the accuracy tends to drop as decision threshold increases. Later, they also investigated the optimisation of generic LAHs \cite{He2014}. The optimisation of LAHs using the GWA is NP-complete, as the branch number of a decision tree is increasing exponentially as the input attribute number increases, and the convergency of GWA directly affects the speed of optimisation. Hence, the evolutionary algorithm takes very long time to optimise LAHs, even if the evolution takes a small number of iterations. When the attribute number is over 60, the time of evolution process on a PC is not acceptable.

Another issue of hierarchical approaches, including generic deep learning, is the lack of linguistic interpretability, especially neural networks have been viewed as a black box. This is because that intermediate variables, being arbitrarily introduced to interconnect the sub-models, do not present any meanings in a real system, and hence it is difficult to give a clear intuitive interpretation \cite{Campello2006}. However, in an LAH, information propagation is completed via the linguistic rules extracted from LDTs. Hence, a transparent hierarchical decision making or classification can be provided.

In a distributed system, we may get partial information from
different information resources, based on which, an initial
estimate/decision could be done locally. Correspondingly, a
collective decision is required. How do these information
resources make contribution to the final collective decision?
In an LAH, an intermediate variable is equivalent to the initial
decision in terms of partial information \cite{He2009a,He2009b}. Hence, the structure of the hierarchy determines the process of decision or fusion process. He et. al. \cite{He2015a} further proposed an off-line optimisation of sensor arrangement for task-oriented sensor fusion, aligning with the structure of an self-organised wireless sensor network.

Mitra et al.\cite{Mitra2002} proposed a fuzzy knowledge-based network, whose structure can be automatically optimised by using linguistic rules of a fuzzy decision tree, thus to help in reducing the complexity of network training for a specific task. An LAH is equivalent to a forward neural network, in which, each neuron is the function of lower layer attributes, but represented by an LDT. The structure of an LAH is different to that of a classic multi-layer feed-forward neural network, in which, all input attributes are in the input layer.  In an LAH, input attributes can feed a neuron (e.g. an LDT) at any layer, but for the simplicity of network structure, they cannot repeatedly feed more than one neuron.

The information propagation from bottom to top of an LAH through the LDTs in the LAH provides a new approach to attribute deep learning. In fact, a neural network or any other machine learning models can be used, instead of an LDT in the LAH. For example, a cascade of CMACs has been proposed in \cite{He2015b}. However, LAH embedded with linguistic decision trees could provide good interpretation for decision making or classification.

He et al. \cite{He2017} used  an LAH to interpret the process of semantic attribute deep learning for spam detection. The LAH was constructed manually in terms of the semantics of attributes. For IoT intelligence or edge computing, the onboard adaptive sensor fusion is a critical challenge, due to the limit of resources in edge devices (e.g. computing capacity, memory and even power supply). Hence, efficiently automatic constructing a linguistic attribute hierarchy becomes necessary for this purpose.

In this research, a self-organised LAH (SOLAH), embedded with LDTs is proposed for the deep learning of attribute semantics in decision making or classification. A distance correlation-based clustering algorithm is proposed to decompose attributes to several clusters, and a linguistic attribute hierarchy is constructed with the produced clusters in terms of the average distance correlation of cluster members to the goal variable. The LDT, fed by a cluster of attributes with lower distance correlation, will be placed in the lower layer of the LAH. The preliminary experiments are conducted on the SMS spam database from UCI machine learning repository \cite{Lichman2013}. A set of databases will be used to validate the performance of the SOLAHs. The experimental results are compared with that of the single LDTs for different databases.
%\vspace{-12pt}
%%%%%%%%%%%%%%%%%%%%%%%%%%%%%%%%%%%%%%%%%%%%%%%%%%%%%%%%%%%%%%%%%%%%%%%%%%%%%%%
%%%%%%%%%%%%%%%%%%%%%%%%%%%%%%%%%%%%%%%%%%%%%%%%%%%%%%%%%%%%%%%%%%%%%%%%%%%%%%%
\section{Existing work in Deep Learning}
Recently, Deep Learning has been put a lot of attention by researchers. It uses multiple processing layers of computational models to learn representations of data with multiple levels of abstraction.  Each successive layer is fed by the outputs of previous layer, forming a hierarchy of attributes from bottom to top.  Deep learning uses the back-propagation algorithm to learn internal parameters that are used to compute the representation in each layer from the representation in the previous layer, thus to find the hidden structure or information in large data sets. Various deep learning architectures such as deep neural networks, convolutional deep neural networks, deep belief networks and recurrent neural networks have been developed. Deep learning has been successfully applied in the areas of speech recognition \cite{Miao2015,ZhangZ2017,Zhang2019}, image processing \cite{HeK2016}, object detection \cite{Girshick2014,Ren2017}, drug discovery \cite{Gawehn2016} and genomics \cite{Park2015}, etc. Especially, deep convolutional nets (ConvNets) have brought about breakthroughs in processing images \cite{Chen2016}, video \cite{Karpathy2014}, speech \cite{Zhang2019} and audio \cite{Lee2009}, whereas recurrent nets have shone light on sequential data such as text and speech \cite{LeCun2015}.

ConvNets are constructed layer by layer for data processing. A classic application of a ConvNet is in image processing, utilising the properties of a colour image, consisting of the RGB channels of pixel values. Since the early 2000s, ConvNets have been successfully applied for the detection, segmentation and recognition of objects and regions based on images, for example, face recognition \cite{Taigman2014}.

Druzhkov and Kustikova \cite{Druzhkov2016} did a survey on deep learning methods for image classification and object detection, covering autoencoders, restricted Boltzmann machines and convolutional neural networks. For example, Szegedy et al. \cite{Szegedy2013} used deep neural networks to solve the problem of object detection for both classifying and  precisely locating objects of various classes; Simonyan et al. \cite{Simonyan2013} used deep fisher networks for the classification of large-scale images from ImageNet (\url{http://image-net.org/}); Krizhevsky et al. \cite{Krizhevsky2012} trained a large deep convolutional neural network to classify the 1.3 million high-resolution images in the LSVRC-2010 ImageNet training set into the 1000 different classes; and  He et al. \cite{HeY2014} proposed an unsupervised feature learning framework, Deep Sparse Coding, which extends sparse coding to a multi-layer architecture for visual object recognition tasks.

Acoustic modeling is another area, where large, deep neural networks have been successfully applied. Like ImageNet, the massive quantities of existing transcribed speech data provide rich resources for deep learning. Dahl \cite{Dahl2015} provided a brief review of the deep neural net approach to large vocabulary speech recognition (LVSR) in his thesis.  Mohamed et al.\cite{Mohamed2012} showed that hybrid acoustic models of pre-trained deep neural networks, instead of Gaussian mixture models (GMMs), could greatly improve the performance of a small-scale phone recognition; By using rectified linear units and dropout, Dahl et al. \cite{Dahl2013} further improve the model for a large vocabulary voice search task.  A combination of a set of deep learning techniques has led to more than 1/3 error rate reduction over the conventional state-of-the-art  GMM-HHM (Hidden Markov Model) framework on many real-world LVCSR tasks \cite{Yu2015}.

The qualitative properties of text data, very different to that of other modality data, provide critical challenges in use of machine learning. Recently Socher et al. \cite{Socher2013} developed recursive deep models for semantic compositionality over a sentiment treebank, and improved the accuracy. Natural language understanding is another exploration of deep learning application, which could make a large impact over the next few years \cite{LeCun2015}.

The property of compositional hierarchies of some signals is
well exploited by deep neural networks through composing
lower-level features to abstract the high-level one. For example,
an image can be represented by a hierarchy from local edges,
motifs, parts, to objects. Similarly, speech and text also have a
hierarchy from sounds to phones, phonemes, syllables, words
and sentences. In other words, this properties promote the capacities
of deep neural networks. However, the interpretability
of the decision making process is still an issue. Also, we cannot
always explicitly see the semantics of higher-level features in
other application domains.

With rapid progress and significant successes in a wide spectrum
of applications, deep learning is being applied in many
safety-critical environments. Ching et al. \cite{Ching2018} forecasted that deep learning enabling changes at both bench and bedside
with the potential to transform several areas of biology and
medicine, although the limited amount of labelled data for
training presents problems as well as legal and privacy constraints
on work with sensitive health records. However, deep
neural networks (DNNs) have been recently found vulnerable
to well-designed input samples called adversarial examples,
and adversarial perturbations are imperceptible to human but
can easily fool DNNs in the testing/deploying stage \cite{Yuan2019}. The
transparency of LAH allows the observation of the decision
process. He et al. \cite{He2017} have investigated the effect of different input attributes on different positions in a linguistic attribute hierarchy. This might indicate a linguistic attribute deep learning could provide good transparency to help people to defend
adversarial perturbations to IoT Intelligence, which is out of
the scope of this research.

GPU has become the necessary hardware facility for the research
on deep learning due to its complexity. Justus et al. \cite{Justus2018}
analysed various factors that influence the complexity of deep
learning, and divided these factors to three categorises: (1)
Layer features, such as activate function, optimiser of weights
(e.g. Gradient Descent) and number of training samples; (2)
layer specific features, including four subcategories: i. Multi-
Layer Perception features (e.g. number of inputs, number
of neurons and number of layers), ii. Convolutional features
(e.g. matrix size, kernel size, stride size, input padding, input
depth and output depth, etc.); iii. Pooling features (e.g. matrix
size, stride size, input padding), iv. Recurrent features (e.g.
Recurrence type, bi-directionality); (3) Hardware Features (e.g.
GPU technology, GPU count, GPU memory, GPU memory
bandwidth, GPU clock speed, GPU cord count, GPU peak
performance, Card Connectivity). The first two categories of
factors determine the performance of the function represented
by a deep learning model, while the GPU feature directly
affect the efficiency of the computing. It is hard to express the
polynomial relationship between the computing cost and all the
various factors, or use a big O(*) to represent the computing
complexity.

%%%%%%%%%%%%%%%%%%%%%%%%%%%%%%%%%%%%%%%%%%%%%%%%%%%%%%%%%%%%%%%%%%%%%%%%%%%%%%%
%%%%%%%%%%%%%%%%%%%%%%%%%%%%%%%%%%%%%%%%%%%%%%%%%%%%%%%%%%%%%%%%%%%%%%%%%%%%%%%
\section{Linguistic attribute deep learning with a LAH}
\subsection{A linguistic attribute hierarchy (LAH)}
The process of aggregation of evidence in multi-attribute decision making or classification based on attributes $x_1,..., x_n$ can be represented with a functional mapping $y = f(x_1,...,x_n)$, where $y$ is the goal variable. However, this mapping is often dynamic and uncertain, and it is difficult to find a mathematic equation to precisely describe the mapping function. An attribute hierarchy represents the function $f$ with a hierarchy of sub-functions, each of which represents a new intermediate attribute. The set of original attributes $\{x_1,..., x_n\}$ is categorised into $m$ clusters $s_1,..., s_m$. When a cluster of attributes is used to make initial decision, an new intermediate attribute is produced by the function of the clustering attributes. One or more intermediate attributes can be combined with another cluster of attributes to make further decision in next level, or multiple intermediate attributes can be directly used to make decision in next level. As there are $m$ attribute clusters, at least $m$ subfunctions are produced.  Namely, $z_i=G_i(s_i)$ for $i=1,...,m$ and $n-1\geq m$, as the maximal level of hierarchy with $n$ input attributes is the cascade hierarchy with $n-1$ subfunctions. The mapping function $f$ is represented by a new function $F$ of the intermediate attributes $z_1,..,z_\tau$ and/or a cluster ($s_i$) of input attributes. The intermediate attributes can be represented by subfunctions of $G_1...G_{\tau}$, fed by lower level of attribute set, which could include intermediate attributes and/or another cluster $s_j$ of input attributes. Hence, $y=f(x_1,...,x_n)$ = $F(z_1,...,z_{\tau}, s_i)$ = $F(G_1(S_1)$,...,$G_{\tau}(S_{\tau}), s_i)$.
%Compared to the original concept of LAH defined in \cite{LJHH08}, the concept of LAH here allows the mixture of intermediate attributes and input attributes %to feed an LDT. Hence, the LAH has more flexible structure.

A linguistic decision tree (LDT) can explicitly model both the uncertainty and vagueness in a mapping system, and linguistic rules extracted from the LDT often implies our knowledge of aggregation. When an LAH uses LDTs to represent the functional mappings between parent and child attribute nodes, the semantic information will be transparently propagated through the hierarchy from bottom to top. Assume the goal variable $y\in \Omega_y$, and a set $\mathcal{L}_y$ of labels can be used to describe the goal variable. An introduced intermediate attribute $z$ will represent an approximate of the goal variable $y$. Namely, $z\in \Omega_y$, and the label set $\mathcal{L}_y$ can be used to
describe $z$. Hence, an LAH can present a hierarchical decision making, and it can provide transparent linguistic interpretation, which helps perform semantic attribute deep learning.

\begin{figure}[htp]
\centering
\subfigure[A cascade LAH]
{\begin{minipage}{0.45\textwidth}
\includegraphics[width=1.78in]{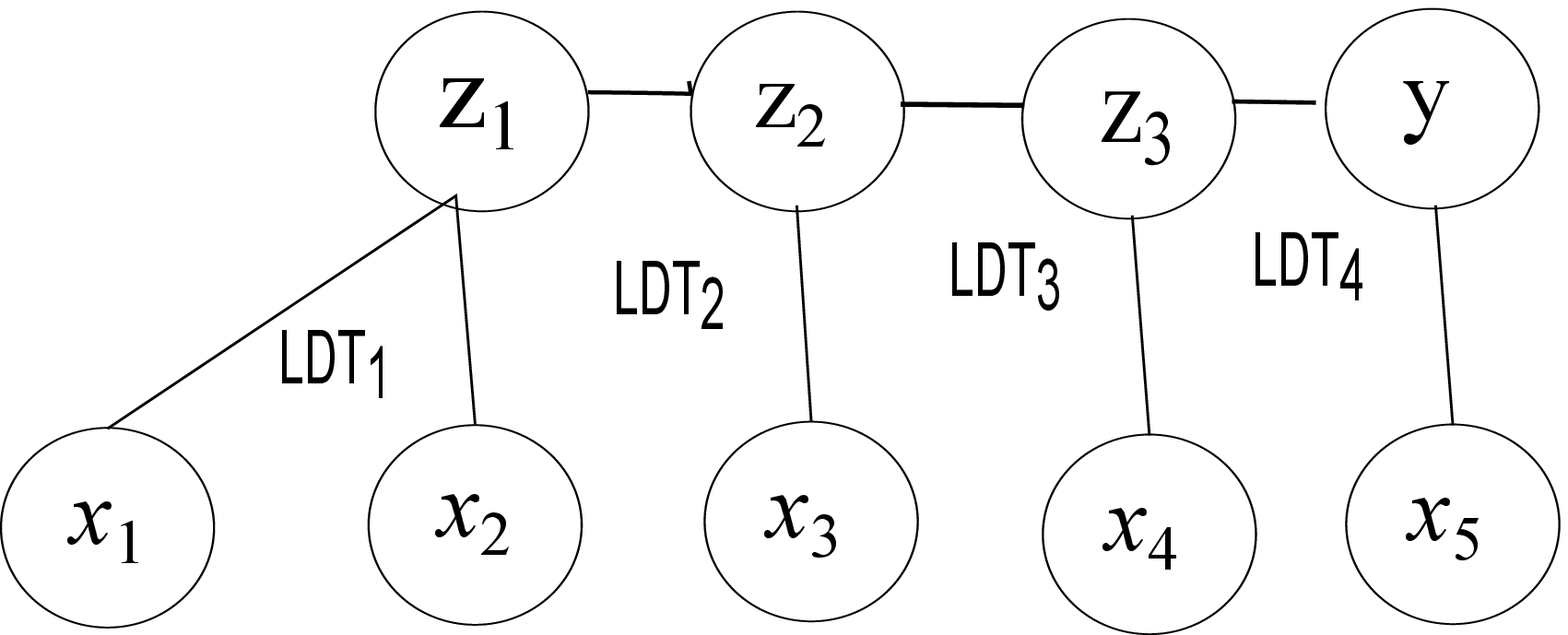}
\end{minipage}}
\subfigure[A general LAH]
{\begin{minipage}{0.45\textwidth}
\includegraphics[width=1.38in]{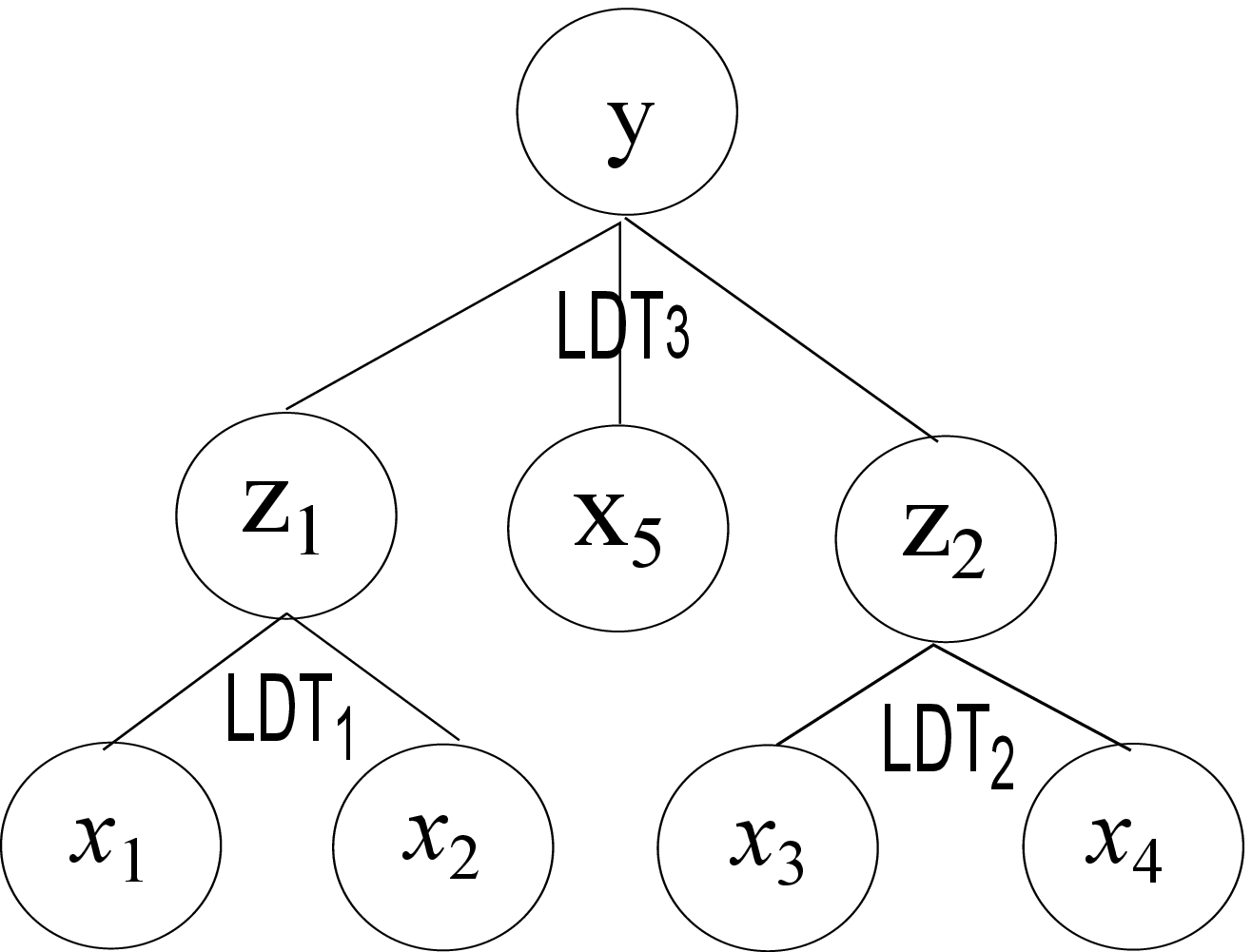}
\end{minipage}}
\caption{Two LAHs with five input attributes} \label{fig:LAH}
\end{figure}

For examples, assume five attributes are fed to an LAH. Fig. \ref{fig:LAH} (a) shows a cascade LAH, embedded with 4 LDTs, through which, the decision information is cascaded from the bottom model $LDT_1$ to the top model $LDT_4$; Fig. \ref{fig:LAH} (b) illustrates a general LAH, embedded with 3 LDTs, through which, the final decision is made by $LDT_3$, based on the two intermediate attributes from $LDT_1$ and $LDT_2$ in the bottom of the LAH, and the additional attribute $x_5$. The intermediate attributes $z_1$ and $z_2$ represent an approximate of the goal variable $y$, $z_1$ is decided by $x_1$ and $x_2$, and $z_2$ is decided by $x_3$ and $x_4$.
%%%%%%%%%%%%%%%%%%%%%%%%%%%%%%%%%%%%%%%%%%%%%%%%%%%%%%%%%%%%%%%%%%%
%%%%%%%%%%%%%%%%%%%%%%%%%%%%%%%%%%%%%%%%%%%%%%%%%%%%%%%%%%%%%%%%%%%

%%%%%%%%%%%%%%%%%%%%%%%%%%%%%%%%%%%%%%%%%%%%%%%%%%%%%%%%%%%%%%%%%%%
%%%%%%%%%%%%%%%%%%%%%%%%%%%%%%%%%%%%%%%%%%%%%%%%%%%%%%%%%%%%%%%%%%%
\subsection{A Linguistic decision tree based on Label Semantics}\label{sec:LDT-LS}
An LDT \cite{Qin2005a,LawryJ2008} is a probabilistic tree under the framework of label semantics \cite{Lawry2006}.  Label semantics introduces two fundamental and interrelated measures: Appropriateness measure ($\mu_L(x)$) and mass assignment ($\nu_x$), where, $\mu_L(x)$ quantifies how a label $L$ is appropriate to describe x based on agent’s knowledge of the current labeling conventions, shown in the collected samples or evidences; $\nu_x$ quantifies how a particular subset of labels is appropriate to describe $x$. The particular subset of all and only the labels that are appropriate to describe $x$ is called focal set, denoted as $F$. Given a set of labels $L = \{s, m, l\}$, the focal set can be $F$ = $\{\{s\}$, $\{s, m\}$, $\{m\}$, $\{m, l\}$, $\{l\}\}$.

An LDT consists of a node set $V$ and an edge set $E$. A node
$v \in V$ is associated to an attribute, and an edge $e\in E$ from
a node is a focal set, which is appropriate to describe the
attribute, indicated by the node. A path from the top node
to a leaf of the LDT is called a branch, denoted as $B$. The
branch length ($l$) is the number of nodes on the branch $B$.
Each branch provides a conjunction of focal sets: $F_1\wedge$ ... $\wedge F_l$, with the inference of a set of mass assignments conditional to the branch $B$, denoted as $\nu_y(F|B)$, for each focal set $F\in \mathcal{F}_y$, where, $\mathcal{F}_y$ is a set of focal sets that are appropriate to describe different values distributed in the domain of the goal variable $y$. The semantics of an LDT is described in Definition \ref{def:linguistic}.
%%%%%%%%%%%%%%%%%%%%%%%%%%%%%%%%%%%%%%%%%%%%%%%%%%%%%%%%%%%%%%%%%%%
%%%%%%%%%%%%%%%%%%%%%%%%%%%%%%%%%%%%%%%%%%%%%%%%%%%%%%%%%%%%%%%%%%%
\begin{definition}[Semantics of an LDT]\label{def:linguistic}
The rule derived from a branch $B_i$ in an LDT is presented as:

\begin{align}\label{eq:rule}
F_{{i}_{1}}\wedge ... \wedge F_{i_l} \rightarrow F:\nu_y(F|B_i),
\end{align}

where, $F_{i_k}$ is the $k$-th focal set on branch $B_i$, and $F_{i_k}\in \mathcal{F}_{x_{i_k}}$. Given attribute values $\vec{x}=(x_1, \ldots,
x_n)$. The mass assignment $\nu_y$ can be obtained using Jeffrey's rule \cite{Jeffrey1990}:

%%%%%%%%%%%%%%%%%%%%%%%%%%%%%%%%%%%%%%%%%%%%%%%%%%%%%%%%%%%%%%%%%%%
%%%%%%%%%%%%%%%%%%%%%%%%%%%%%%%%%%%%%%%%%%%%%%%%%%%%%%%%%%%%%%%%%%%
\begin{gather}\label{eq:Jeffery}
\nu_y(F)=\sum_{i=1}^{\beta} \mu_{B_i}(\vec{x})\nu_y(F|B_i), F\in \mathcal{F}_y
\end{gather}
%%%%%%%%%%%%%%%%%%%%%%%%%%%%%%%%%%%%%%%%%%%%%%%%%%%%%%%%%%%%%%%%%%%
%%%%%%%%%%%%%%%%%%%%%%%%%SECTION%%%%%%%%%%%%%%%%%%%%%%%%%%%%%%%%%%%
where $\beta$ is the branch number of the LDT, and $\nu_y(F|B_i)$ is equivalent to the conditional probability $p(F|B_i)$. $x_{i_j}$ denotes the $j$-th attribute on the $i$-th branch. Assume
a focal set $F_{i_j}$ is appropriate to describe attribute $x_{i_j}$ on the branch $B_i$. The appropriateness measure of $\vec{x}$ on branch $B_i$ is the product of all mass assignments on focal sets that are appropriate to describe the corresponding nodes on the branch, as expressed in Formula (\ref{eq:pr-branch}).
%%%%%%%%%%%%%%%%%%%%%%%%%%%%%%%%%%%%%%%%%%%%%%%%%%%%%%%%%%%%%%%%%%%
%%%%%%%%%%%%%%%%%%%%%%%%%%%%%%%%%%%%%%%%%%%%%%%%%%%%%%%%%%%%%%%%%%%
\begin{gather}\label{eq:pr-branch}
\mu_{B_i}(\vec{x}) = \prod_{j=1}^l \nu_{x_{i_j}}(F_{i_j}).
\end{gather}
%%%%%%%%%%%%%%%%%%%%%%%%%%%%%%%%%%%%%%%%%%%%%%%%%%%%%%%%%%%%%%%%%%%
%%%%%%%%%%%%%%%%%%%%%%%%%%%%%%%%%%%%%%%%%%%%%%%%%%%%%%%%%%%%%%%%%%%
\end{definition}
%%%%%%%%%%%%%%%%%%%%%%%%%%%%%%%%%%%%%%%%%%%%%%%%%%%%%%%%%%%%%%%%%%%
%%%%%%%%%%%%%%%%%%%%%%%%%%%%%%%%%%%%%%%%%%%%%%%%%%%%%%%%%%%%%%%%%%%
%%%%%%%%%%%%%%%%%%%%%%%%%%%%%%%%%%%%%%%%%%%%%%%%%%%%%%%%%%%%%%%%%%%
%%%%%%%%%%%%%%%%%%%%%%%%%%%%%%%%%%%%%%%%%%%%%%%%%%%%%%%%%%%%%%%%%%%
\section{The semantics of a linguistic attribute hierarchy}
When an LAH is represented a hierarchy of LDTs, the information is propagated through the LDTs from information sources (e.g.
sensors) in the bottom to the decision variable on the top. Therefore, the rules can be derived as conditional expressions through Formula \ref{eq:rule} in the label semantics framework. The output variable (either an intermediate attribute or a goal variable) of each LDT in the LAH can be calculated with Formula (\ref{eq:Jeffery}).
This provides an approach to quantifying the degree of our belief how each focal set $F\in \mathcal{F}_y$ is appropriate to describe the goal,
given partial input attributes and/or the previous results of decision or classification (intermediate attributes) in lower level.

Now, we use the LAH in Fig. \ref{fig:LAH} (b) as an example to demonstrate the upwards information propagation through an LAH. In Fig. \ref{fig:LAH} (b), two LDTs ($LDT_1$, $LDT_2$) are located in the bottom of the LAH. $y=f(z_1,z_2, x_5)= f(g_1(x_1,x_2),g_2(x_3,x_4),x_5)$. The
mappings $g_1$, $g_2$, and $f$ are represented in the form of  linguistic decision trees (i.e. $LDT_1$, $LDT_2$ and $LDT_3$). $LDT_1$ provides the function of quantifying our belief of $z_1$ on goal labels in terms of input attributes $x_1$ and $x_2$, $LDT_2$ presents the function of quantifying our belief of $z_2$ on goal labels in terms of input attributes $x_3$ and $x_4$, and $LDT_3$ offers the function of quantifying our final belief of the goal variable $y$ on its labels
in terms of lower level believes of $z_1$ and $z_2$, as well as another input attribute $x_5$ from information source (e.g. sensors).
%%%%%%%%%%%%%%%%%%%%%%%%%%%%%%%%%%%%%%%%%%%%%%%%%%%%%%%%%%%%%%%%%%%
%%%%%%%%%%%%%%%%%%%%%%%%%%%%%%%%%%%%%%%%%%%%%%%%%%%%%%%%%%%%%%%%%%%
Assume the input attributes are clustered to $\kappa$ subsets ${s_1,...s_\kappa}$, which can feed the intermediate
 attributes or/and the goal variable in an LAH. As an intermediate attribute $z_i \in \Omega_y$ is the approximate of the goal variable $y$, it can be estimated by a decision making model (e.g. LDT), fed with $s_i$. Hence, the decision making model can be trained with the samples of ($S_i,y$). Given the values of all input attributes $\vec{x}=(x_1, \ldots, x_n)$, the mass assignments of all intermediate attributes and the goal variable can be estimated with Formula (\ref{eq:Jeffery}) through all the LDTs in the LAH. Namely, the decision information is propagated through the bottom LDTs to the top LDT in the LAH. The semantics of an LAH is defined as Definition \ref{def:HLinguistics}.

\begin{definition}[Semantics of an LAH]\label{def:HLinguistics}
  The semantics of an LAH is the synthetisation of rules extracted from the branches in the LAH, which are allocated by the given sample, $\vec{x}=(x_1, \ldots, x_n)$. Assume $k$ outputs (i.e. $k$ intermediate attributes) of LDT($t_1$) ... LDT($t_k$) are the inputs of LDT($t_i$), the rule will be $B^{t_1} \wedge ... \wedge B^{t_k} \rightarrow B^{t_i}$, where $B^{t_1}$ ... $B^{t_k}$ can be derived in the form of Formula (\ref{eq:rule}), by an LDT, either directly based on  information sources (i.e. the input attributes), or based on the intermediate attributes from the lower level of LDTs.
\end{definition}

%%%%%%%%%%%%%%%%%%%%%%%%%%%%%%%%%%%%%%%%%%%%%%%%%%%%%%%%%%%%%%%%%%%
%%%%%%%%%%%%%%%%%%%%%%%%%%%%%%%%%%%%%%%%%%%%%%%%%%%%%%%%%%%%%%%%%%%
For the instance of Fig. \ref{fig:LAH} (b), given all the input attribute values,
$\vec{x}$ = $\{x_1$, $x_2$, $x_3$, $x_4$, $x_5\}$ in all samples. For generality, we use $\ell$ to denote the label expression, associated to an edge in an LDT. For the special case, the edges of an LDT are associated to a focal set, $\ell$ represents a focal set $F$.  Hence, the semantics of each decision tree can be described as:

For $LDT_1(x_1,x_2)$, as it is fed with two attributes, the maximum branch length is 2. Hence, the rule corresponding to the branch $B^1_{i}$
can be:
\begin{align}
\ell^1_{{i}_{1}}\wedge \ell^1_{{i}_2} \rightarrow
F:\nu_y(F|B^1_{i}),  F\in \mathcal{F}_y.\nonumber
\end{align}

Similarly, for $LDT_2(x_3,x_4)$, the rule corresponding to the branch $B^2_{j}$
can be:
\begin{align}
\ell^2_{{j}_1}\wedge \ell^2_{{j}_2} \rightarrow
F:\nu_y(F|B^2_{j}), F\in \mathcal{F}_y.\nonumber
\end{align}

For $LDT_3(z_1,z_2, x_5)$, the maximum branch length is 3. Hence, the rule corresponding to the branch $B^3_{k}$
can be:
\begin{align}
\ell^3_{{k}_1}\wedge \ell^3_{{k}_2} \wedge \ell^3_{{k}_3} \rightarrow
F:\nu_y(F|B^3_{k}), F\in \mathcal{F}_y.\nonumber
\end{align}

The synthetic semantics of the LAH is:
\begin{align}
(\ell^1_{{i}_{1}}\wedge \ell^1_{{i}_2}) \vee (\ell^2_{{j}_{1}}\wedge \ell^2_{{j}_2})& \rightarrow \ell^3_{{k}_1}\wedge \ell^3_{{k}_2}\wedge \ell^3_{{k}_3} \nonumber\\
&\rightarrow F:\nu_y(F|B^3_{k}), F\in \mathcal{F}_y.\nonumber
\end{align}

%%%%%%%%%%%%%%%%%%%%%%%%%%%%%%%%%%%%%%%%%%%%%%%%%%%%%%%%%%%%%%%%%%%
%%%%%%%%%%%%%%%%%%%%%%%%%%%%%%%%%%%%%%%%%%%%%%%%%%%%%%%%%%%%%%%%%%%
%%%%%%%%%%%%%%%%%%%%%%%%%%%%%%%%%%%%%%%%%%%%%%%%%%%%%%%%%%%%%%%%%%%
%%%%%%%%%%%%%%%%%%%%%%%%%%%%%%%%%%%%%%%%%%%%%%%%%%%%%%%%%%%%%%%%%%%
\section{Construction of Linguistic Attribute Hierarchy}
%%%%%%%%%%%%%%%%%%%%%%%%%%%%%%%%%%%%%%%%%%%%%%%%%%%%%%%%%%%%%%%%%%%%
%%%%%%%%%%%%%%%%%%%%%%%%%%%%%%%%%%%%%%%%%%%%%%%%%%%%%%%%%%%%%%%%%%%%
The basic idea for the construction of a Linguistic Attribute Hierarchy includes two steps:
(1) Decomposition of attributes with a distance correlation based clustering algorithm;\\
(2) Self-organisation of a linguistic attribute hierarchy in terms of the distance correlation between clustered attributes and the goal variable.
%%%%%%%%%%%%%%%%%%%%%%%%%%%%%%%%%%%%%%%%%%%%%%%%%%%%%%%%%%%%%%%%%%%%
%%%%%%%%%%%%%%%%%%%%%%%%%%%%%%%%%%%%%%%%%%%%%%%%%%%%%%%%%%%%%%%%%%%%
\subsection{Attribute decomposition based on their distance correlation}
%%%%%%%%%%%%%%%%%%%%%%%%%%%%%%%%%%%%%%%%%%%%%%%%%%%%%%%%%%%%%%%%%%%%
%%%%%%%%%%%%%%%%%%%%%%%%%%%%%%%%%%%%%%%%%%%%%%%%%%%%%%%%%%%%%%%%%%%%
\subsubsection{Distance correlation}
Distance correlation can be used to statistically measure the dependence between two random variables or vectors, which could have different dimensions. It is zero if and only if the random variables are statistically independent. Assume $(x_i, y_i), i= 1, 2, ..., n$ be a set of samples. The $n$ by $n$ distance matrices $(a_{i,j} )$ and $(b_{i,j})$ present the all pairwise distances for $x$ and $y$ in all the samples, respectively.
%%%%%%%%%%%%%%%%%%%%%%%%%%%%%%%%%%%%%%%%%%%%%%%%%%%%%%%%%%%%%%%%%%%%
%%%%%%%%%%%%%%%%%%%%%%%%%%%%%%%%%%%%%%%%%%%%%%%%%%%%%%%%%%%%%%%%%%%%
\begin{align}
a_{i,j} = \parallel  x_{i}-x_{j}\parallel;
b_{i,j} = \parallel y_{i}-y_{j}\parallel.
\end{align}
%%%%%%%%%%%%%%%%%%%%%%%%%%%%%%%%%%%%%%%%%%%%%%%%%%%%%%%%%%%%%%%%%%%%
%%%%%%%%%%%%%%%%%%%%%%%%%%%%%%%%%%%%%%%%%%%%%%%%%%%%%%%%%%%%%%%%%%%%
where $|| \bullet ||$ denotes Euclidean norm. Then take all doubly centered distances:
%%%%%%%%%%%%%%%%%%%%%%%%%%%%%%%%%%%%%%%%%%%%%%%%%%%%%%%%%%%%%%%%%%%%
%%%%%%%%%%%%%%%%%%%%%%%%%%%%%%%%%%%%%%%%%%%%%%%%%%%%%%%%%%%%%%%%%%%%
\begin{align}
A_{i,j} = a_{i,j}-\bar{a}_{i.}-\bar{a}_{.j}+\bar{a}_{..},
B_{i,j} = b_{i,j}-\bar{b}_{i.}-\bar{b}_{.j}+\bar{b}_{..},
\end{align}
%%%%%%%%%%%%%%%%%%%%%%%%%%%%%%%%%%%%%%%%%%%%%%%%%%%%%%%%%%%%%%%%%%%%
%%%%%%%%%%%%%%%%%%%%%%%%%%%%%%%%%%%%%%%%%%%%%%%%%%%%%%%%%%%%%%%%%%%%
where, $\bar{a}_{i.}$ is the $i$-th row mean, $\bar{a}_{.j}$ is the $j$-th column mean, and $\bar{a}_{..}$ is the grand mean of the distance matrix for the $x$ samples. The notation is similar for the $b$ values on the $y$ samples. The squared sample distance covariance (a scalar) is simply the arithmetic average of the products $A_{i,j}$ and $B_{i,j}$:
%%%%%%%%%%%%%%%%%%%%%%%%%%%%%%%%%%%%%%%%%%%%%%%%%%%%%%%%%%%%%%%%%%%%
%%%%%%%%%%%%%%%%%%%%%%%%%%%%%%%%%%%%%%%%%%%%%%%%%%%%%%%%%%%%%%%%%%%%
\begin{align}
dCov_n^2 (x,y)=\frac{1}{n^2}\sum_{i=1}^n\sum_{j=1}^n A_{i,j}B_{i,j}.
\end{align}
%%%%%%%%%%%%%%%%%%%%%%%%%%%%%%%%%%%%%%%%%%%%%%%%%%%%%%%%%%%%%%%%%%%%
%%%%%%%%%%%%%%%%%%%%%%%%%%%%%%%%%%%%%%%%%%%%%%%%%%%%%%%%%%%%%%%%%%%%
The sample distance variance is the square root of
\begin{align}
  dVar_{n}^{2}(x)= dCov_{n}^{2}(x,x) = \frac {1}{n^{2}}\sum _{i,j}A_{i,j}^{2},
\end{align}
%%%%%%%%%%%%%%%%%%%%%%%%%%%%%%%%%%%%%%%%%%%%%%%%%%%%%%%%%%%%%%%%%%%%
%%%%%%%%%%%%%%%%%%%%%%%%%%%%%%%%%%%%%%%%%%%%%%%%%%%%%%%%%%%%%%%%%%%%
The distance correlation  of two random variables can be calculated through dividing their distance covariance by the product of their distance standard deviations, as Fomula (\ref{eq:crown}).
\begin{align}\label{eq:crown}
dCorr(x,y)=\frac {dCov(X,Y)}{\sqrt {dVar(x)\ dVar(y)}}.
\end{align}
%%%%%%%%%%%%%%%%%%%%%%%%%%%%%%%%%%%%%%%%%%%%%%%%%%%%%%%%%%%%%%%%%%%%
%%%%%%%%%%%%%%%%%%%%%%%%%%%%%%%%%%%%%%%%%%%%%%%%%%%%%%%%%%%%%%%%%%%%
Distance correlation has the properties:
\begin{itemize}
 \item[(1)] $0\leq dCorr_{n}(x,y)\leq 1$, $0\leq dCorr(x,y)\leq 1$;
 \item[(2)] The distance correlation matrix is symmetric (i.e. $dCorr(i,j) = dCorr(j,i)$), and $dCorr(i,i)$=1.
 \item[(3)] $dCorr(x,y)=0$, if and only if $x$ and $y$ are independent;
 \item[(4)] $dCorr_{n}(x,y)=1$ and $dCorr(x,y)=1$ indicates that the linear subspaces spanned by $x$ and $y$ samples respectively almost surely have an identical dimension \cite{Szekely2009}.
\end{itemize}
%%%%%%%%%%%%%%%%%%%%%%%%%%%%%%%%%%%%%%%%%%%%%%%%%%%%%%%%%%%%%%%%%%%%
%%%%%%%%%%%%%%%%%%%%%%%%%%%%%%%%%%%%%%%%%%%%%%%%%%%%%%%%%%%%%%%%%%%%
%%%%%%%%%%%%%%%%%%%%%%%%%%%%%%%%%%%%%%%%%%%%%%%%%%%%%%%%%%%%%%%%%%%%
\subsubsection{Clustering attributes based on distance correlation}
As we do not need to consider the distance correlation between an attribute and itself, we set the diagonal of the distance correlation matrix to zeros for the convenience of computing. First, we find the maximum value in the distance correlation matrix $dCorr$. The maximum value is denoted as $d_{max}$, the column of the maximum value in the matrix is denoted as $i_{max}$. Here we set a value $\alpha$ as the range of distance correlation difference in a cluster. Namely, the distance correlation of attributes in a cluster, correlating to attribute $i_{max}$ will be in $(d_{max}-\alpha, d_{max}]$. A feasible approach to setting the value of $\alpha$ is:
\begin{align}
 (max(dCorr)-min(dCorr))/k,
\end{align}

 Usually, $min(dCorr)$  is not a zero, and $k$ is the preset number of clusters.  Secondly, all relevant columns and rows, where the identified cluster members in the cluster are located, are set to zeros. The process is repeated with starting to  find next attribute with largest distance correlation in the unvisited attributes for the next cluster, until all elements of $dCorr$ are zeros. Algorithm \ref{alg:dCorrClustering} provides the pseudo-code of the clustering algorithm, where the produced clusters are saved in $\mathcal{S}$.
\begin{algorithm}[ht]
\caption{Distance Correlation Clustering ($dCorr$,$k$)}\label{alg:dCorrClustering}
\footnotesize
\begin{algorithmic}[1]
\STATE Initialise($\mathcal{S})$;
\STATE t=0;
\WHILE {($dCorr \neq [0]$)}
    \STATE $t$ = $t$+1;
    \STATE $\alpha=(max(dCorr)-min(dCorr))/k$;
    \STATE $[d_{max},i_{max}]=max(dCorr)$;
    \STATE $T \leftarrow i_{max}$;
    \STATE $T \leftarrow find\_i(d_{max}> dCorr_{i,i_{max}}\geq d_{max}-\alpha)$);
    \STATE $S_t \leftarrow T$;
    \STATE $clearCols(dCorr, T$);
    \STATE $clearRows(dCorr, T$);
\ENDWHILE
\end{algorithmic}
\end{algorithm}

\subsection{Self-organisation of linguistic attribute hierarchy}
Once attributes are clustered based on their distance correlation, we can construct a linguistic attribute hierarchy in terms of  the distance correlation between clustered attributes and the goal variable. If an attribute has stronger distance correlation with decision variable, this indicates the attribute has stronger linear correlation to the goal variable, namely, the mapping function between the attribute and the goal could be more linear. In other words, it is easier to make decision in terms of the attribute. Our preliminary experimental results show that attributes, which have stronger distance correlation to the goal variable, should be fed to the LDT in higher layer of  the linguistic attribute hierarchy. The average of distance correlation between attributes in a cluster and the goal variable can be calculated with Formula (\ref{eq:average-dCorr}):
%%%%%%%%%%%%%%%%%%%%%%%%%%%%%%%%%%%%%%%%%%%%%%%%%%%%%%%%%%%%%%%%%%%%
%%%%%%%%%%%%%%%%%%%%%%%%%%%%%%%%%%%%%%%%%%%%%%%%%%%%%%%%%%%%%%%%%%%%
 \begin{align}\label{eq:average-dCorr}
 \overline{dCorr}(s) = \sum_{x\in s} dCorr(x,y)/t, t=|s|.
 \end{align}
%%%%%%%%%%%%%%%%%%%%%%%%%%%%%%%%%%%%%%%%%%%%%%%%%%%%%%%%%%%%%%%%%%%%
%%%%%%%%%%%%%%%%%%%%%%%%%%%%%%%%%%%%%%%%%%%%%%%%%%%%%%%%%%%%%%%%%%%%
 where, $x$ is the attributes in cluster $s$, $y$ is the goal variable, and $t$ is the size of cluster (i.e. number of attributes in cluster $s$).

 The basic idea of constructing an LAH based on the produced clusters is that:
 \begin{itemize}
 \item[(1)] calculating average distance correlation for each produced cluster;
 \item[(2)] Sort all average distance correlations between clusters and decision variable;
 \item[(3)] The LDT fed by the cluster with lower average distance correlation will be at lower level of the LAH;
 \item[(4)] The LDTs fed by the clusters that have similar average values of distance correlation will be at the same level of the LAH. A threshold $\theta$ is set for the assessment of clusters at the same level. The outputs of all LDTs at the same level will be the partial inputs of the next LDT in the next level of LAH;
 \item[(5)] The LDT fed by the cluster with the highest average distance correlation will be on the top of the LAH.
\end{itemize}

     %The top LDT is with the highest F-score. LDTs that have similar F-scores will stay on the same layer, namely,  $(F(LDT_i)-F(LDT_j))<\theta$. The output of these LDTs will be the input of next level of LDT. If there are more %than two LDTs in one layer, their outputs can be inputs of  a new LDT, and then the output of the new LDT will be the input of an LDT in next layer. Fig. \ref{fig:SOLAH} illustrates the process of the construction of an %LAH. LDT5 obtains F-score 0.80, LDT2, LDT3 and LDT4 obtain F-scores 0.745, 0.75 and 0.753, respectively, the values of which are very close, and LDT1 obtains lowest F-score, 0.70. Therefore, LDT5 is placed on the top of the %LAH, LDT2, LDT3, and LDT4 are placed at the same layer, and we insert a new LDT (i.e. LDT6) as a layer to make decision fusion of the three LDTs. The output of $z_6$ is with the subset of attributes for LDT5. LDT1 is at the %bottom of the LAH. The output $z_1$ of LDT1 is with the subset of attributes for LDT2.

Assume $\overline{dCorr}(s_i)$ represents the average distance correlation of cluster $s_i$ to the goal variable, and $\theta$ represents the maximum difference of distance correlation values of clusters at the same level to the goal variable. Namely, if
%%%%%%%%%%%%%%%%%%%%%%%%%%%%%%%%%%%%%%%%%%%%%%%%%%%%%%%%%%%%%%%%%%%%
%%%%%%%%%%%%%%%%%%%%%%%%%%%%%%%%%%%%%%%%%%%%%%%%%%%%%%%%%%%%%%%%%%%%
\begin{align}
|\overline{dCorr}(s_i)- \overline{dCorr}(s_j)|<\theta,
\end{align}
%%%%%%%%%%%%%%%%%%%%%%%%%%%%%%%%%%%%%%%%%%%%%%%%%%%%%%%%%%%%%%%%%%%%
%%%%%%%%%%%%%%%%%%%%%%%%%%%%%%%%%%%%%%%%%%%%%%%%%%%%%%%%%%%%%%%%%%%%
then the LDTs fed with $s_i$ and $s_j$ will be at the same level in the LAH.

Algorithm \ref{alg:SOLAH} provides the pseudo-code of the self-organisation of LAH, where $dCorrXY$ is the vector of distance correlation of all attributes to the goal variable, $\mathcal{S}$ is the set of clusters produced by the distance correlation-based clustering algorithm above, $\mathcal{R}$ is the set of all the average distance correlations for cluster set $\mathcal{S}$ with $K$ clusters to the goal variable, $Z$ is the index of an intermediate attribute, which starts from $n+1$, the input attribute number,  $\mathcal{I}$ is used to save the indices of intermediate attributes and the corresponding clusters at the same level, $\mathcal{H}$ is used to save the hierarchy to be constructed. For all clusters, if $R_i-R_a <\theta$, save the intermediate attribute index $Z$ to $\mathcal{I}$, and append $Z$ and the cluster $S_{t_i}$ to the hierarchy $\mathcal{H}$, else, start a new level, append current $Z$, other intermediate attributes saved in $\mathcal{I}$ and corresponding cluster $S_{t_i}$ to the hierarchy $\mathcal{H}$, and save current $Z$ to $\mathcal{I}$. If more than one LDTs at the same level in the LAH, and there is no further cluster to be constructed, then the outputs of those LDTs will be the inputs of the top LDT in the LAH. If the LDT constructed with the last cluster does not share the same level with other LDTs, then it will be the top LDT of LAH.
\begin{algorithm}[ht]
\caption{SOLAH($dCorrXY$, $\mathcal{S}$, $\theta$)}\label{alg:SOLAH}
\footnotesize
\begin{algorithmic}[1]
\STATE [$\mathcal{R}$] = avDistCorr($dCorrXY,\mathcal{S}$);
\STATE $Z = n$; $K$=|$\mathcal{S}$|;
\STATE [$R,t$] = sort($\mathcal{R}$); \%increasingly
\STATE $\mathcal{I} = \phi$;$\mathcal{H} = \phi$;
\STATE $i$ = 0, $a$ = 0;
\STATE $H_i = \phi$;
\WHILE {($i < K$)}
    \STATE $i=i$+1; $Z = Z$+1,
    \IF {(($R_i-R_a<\theta$)}
        \STATE $\mathcal{I} \leftarrow(Z)$;
        \STATE $\mathcal{H} = \leftarrow (Z, \mathcal{S}_{t_i}$);
    \ELSE
        \STATE $a = i$; \% starting a new level
        \STATE $\mathcal{H} \leftarrow(Z, \mathcal{I}, \mathcal{S}_{t_i}$);
        \STATE $\mathcal{I}=Z$;
    \ENDIF
\ENDWHILE
\IF {($|\mathcal{I}|$ > 1)}
    \STATE $H \leftarrow(Z+1,I)$;
    \STATE $Y = Z$+1;
\ELSE
    \STATE $Y = Z$;
\ENDIF
\end{algorithmic}
\end{algorithm}

%\begin{figure}[ht!]
%\centering
%\subfigure[Individual LDTs]
%{\begin{minipage}{0.45\textwidth}
%\includegraphics[width=3.2in]{LDTs.eps}
%\end{minipage}}
%\subfigure[A self-organised LAH ]
%{\begin{minipage}{0.45\textwidth}
%\includegraphics[width=2.8in]{SOLAH.eps}
%\end{minipage}}
%\subfigure[A simplified LAH ]
%{\begin{minipage}{0.45\textwidth}
%\includegraphics[width=2.8in]{SOLAH-1.eps}
%\end{minipage}}
%\caption{An example of Self-Organised LAH} \label{fig:SOLAH}
%\end{figure}
\subsection{Training of a linguistic attribute hierarchy}
In 2009, He and Lawry \cite{He2009a,He2009b} first time proposed using the domain and labels of the goal variable to describe intermediate attributes. This made the hierarchy training became possible, and the hierarchical decision making became meaningful. In \cite{He2014}, He and Lawry developed a non-recursively post-order traversal algorithm to train a given LAH, where all LDTs are trained with LID3 in a bottom-up
way. Intermediate attribute values are estimated with the trained LDTs, and then they are input to next level LDTs.
For example, in Fig. \ref{fig:LAH} (b), $LDT_1$, $LDT_2$ and $LDT_3$ are trained in turn, where, the intermediate variables $z_1$ and $z_2$ use the corresponding $y$ values in all training points, as the goal values of $LDT_1$ and $LDT_2$ respectively. After $LDT_1$ and $LDT_2$ are trained, the intermediate variables $z_1$ and $z_2$ for the training samples can be estimated  by $LDT_1$ and $LDT_2$, respectively.
The estimated values of $z_1$ and $z_2$, the values of $x_5$, and the values of the goal variable $y$ in the training samples are used to train the $LDT_3$, and finally the goal variable $y$ can be estimated by $LDT_3$. Here, we introduce a recursive postorder implementation of the LAH training algorithm (see Algorithm \ref{alg:LAH-Train}).

%\begin{table}[ht!]
%\centering
%\caption{Symbol Description in Algorithm \ref{alg:LAH-Train}} \label{tab:symbols}
%\scriptsize
%\begin{tabular}{l|p{148pt}}
%\hline
%Symbols & Description\\
%\hline\hline
%$\mathcal{H}$       & An LAH to be trained, $\mathcal{D}$ is the data set, namely, all samples of $(\vec{x},y)$\\
%$v$                 & The node $v  \in \mathcal{H}$\\
%$v.c_i$             & The $i$-th child of node $v$\\
%$k$                 & The number of child nodes of node $v$\\
%$s_v$               & The attribute set that feeds the $LDT_v$, including either intermediate attributes or original input attributes\\
%$\mathcal{D}_{s_v}$ & Represent the values of attribute $s_v$ in the data samples $\mathcal{D}$, which may includes the values of intermediate attributes estimated by the LDTs in lower levels\\
%\hline
%\end{tabular}
%\end{table}
$\mathcal{H}$ is a global variable, indicating the LAH to be trained. Initially, $v$ is the root of the LAH to be trained, and all leaves in the LAH are tagged as visited, as they are the input attributes from sensor nodes. $v.ch$ represents the child set of node $v$, indicating the attributes that feed to the node $v$. $S$ is used to save the attributes that are not on the branch from the top root to the current node $v$, and $S\setminus v$ is to remove $v$ from set $S$.
%%%%%%%%%%%%%%%%%%%%%%%%%%%%%%%%%%%%%%%%%%%%%%%%%%%%%%%%%%%%%%%%%%%
%%%%%%%%%%%%%%%%%%%%%%%%%%%%%%%%%%%%%%%%%%%%%%%%%%%%%%%%%%%%%%%%%%%
\begin{algorithm}[htbp]
\caption{postorder($v$,$S$)}\label{alg:LAH-Train}
\footnotesize
\begin{algorithmic}
   \IF {(node $v\in \mathcal{H}$ has been visited)}
       \STATE return;
   \ELSE
      \FOR {($i$=1..$k$)}
        \STATE $postorder$($v.ch_i$, $S\setminus v$);
      \ENDFOR
      \STATE $LDT_v$=$LID3(v, S)$;
      \STATE $z_v=LDT_v(\mathcal{D})$;
      \STATE $\mathcal{D} \leftarrow z_v$;
   \ENDIF
\end{algorithmic}
\end{algorithm}

%%%%%%%%%%%%%%%%%%%%%%%%%%%%%%%%%%%%%%%%%%%%%%%%%%%%%%%%%%%%%%%%%%%
%%%%%%%%%%%%%%%%%%%%%%%%%%%%%%%%%%%%%%%%%%%%%%%%%%%%%%%%%%%%%%%%%%%
\subsection{The LID3 algorithm for LDT training}
LID3 was used to train an LDT with a given database \cite{Qin2005a,He2009a}. It is an update of the classic ID3
algorithm \cite{Quinlan1986}, through combining label semantics.  The training process is conducted through selecting the attribute that obtains the maximum information gain to extend a branch.
The functions used in LID3 have been formulated in \cite{Qin2005a,He2009a}, such as information entropy of a branch ($E$), expected entropy ($EE$) when an attribute node $x$ is added to a branch, and information gain (IG). Here, a recursive algorithm implementation of LID3 is proposed (Algorithm \ref{alg:LID3}).
%\begin{table}[ht!]
%\centering
%\caption{Symbol Description in Algorithm \ref{alg:LAH-Train}} \label{tab:LID3-symbols}
%\scriptsize
%\begin{tabular}{l|p{148pt}}
%\hline
%Symbols & Description\\
%\hline\hline
%$\mathcal{T}$       & A global variable of an LDT\\
%$\mathcal{D}$       & A global variable of the training data set, ($\vec{x},y)$\\
%$S$                 & An attribute set, initially, $S$ is the full set of input attributes\\
%$\{\cdot\}$         & A set that contains element `$\cdot$'.\\
%$B_v$               & A branch, from the root to node $v$  \\
%$C$                 & A class of the goal variable \\
%$t$                 & Probability threshold \\
%$E$                 & The entropy of a branch B\\
%$EE$                & The expected entropy, when a focal set edge is appended to a branch\\
%$IG$                & The information gain, when an attribute is added to a branch\\
%$P(C|B)$            & The probability of that the object belongs to $C$, given the condition of a branch $B$\\
%\hline
%\end{tabular}
%\end{table}

A threshold $\vartheta$ is setup to stop the branch extension when the conditional probability $P(C|B)$ reaches the threshold $\vartheta$. The maximum of branch length is the number of attributes that feed the LDT. Initially, $T = \phi$ (empty), hence, current node point $B_v = \phi$ as well. For each branch, the conditional probabilities for all classes $C\in \mathcal{C}$ are calculated. The most informative attribute is selected as the next node to be extended to the existing LDT, and all focal sets that are appropriated to describe the attribute in the domain of $\Omega_x$ will be appended to the tree $T$. The process is continued until the maximum conditional probability arrives the specified threshold.
\begin{algorithm}[ht]
\caption{$LID3(B_v, S$)}\label{alg:LID3}
\footnotesize
\begin{algorithmic}
\FOR {(all $C_i\in \mathcal{C}_y$)}
     \STATE $P(C_i|B_v)$ = conditionProb($B_v$);
\ENDFOR
\IF {($max(\{P\})\geq \vartheta$) or $S = \phi$}
     \STATE return;
\ENDIF
\FOR {(all $x \in S$)}
    \STATE $IG(B_v,x) = E(B_v)-EE(B_v, x)$;
\ENDFOR
\STATE $\hat{x} = argmax_{x\in S}(\{IG\})$;
\STATE $B_{\hat{v}} = B_v+x$
\STATE $T \leftarrow {F\in \mathcal{F}_{\hat{x}}}$;
\STATE $S = S\setminus \hat{x}$;
\FOR {(all $F\in \mathcal{F}_{\hat{x}}$)}
    \STATE $LID3 (B_{\hat{v}}, S)$;
\ENDFOR
\end{algorithmic}
\end{algorithm}
%%%%%%%%%%%%%%%%%%%%%%%%%%%%%%%%%%%%%%%%%%%%%%%%%%%%%%%%%%%%%%%%%%%
%%%%%%%%%%%%%%%%%%%%%%%%%%%%%%%%%%%%%%%%%%%%%%%%%%%%%%%%%%%%%%%%%%%
%%%%%%%%%%%%%%%%%%%%%%%%%%%%%%%%%%%%%%%%%%%%%%%%%%%%%%%%%%%%%%%%%%%
%%%%%%%%%%%%%%%%%%%%%%%%%%%%%%%%%%%%%%%%%%%%%%%%%%%%%%%%%%%%%%%%%%%
\section{Experiments and Evaluation}
\subsection{Experiment methodologies}
All databases, used for the experiments, are from UCI machine learning repository \cite{Lichman2013}. The experiments are conducted in two stages:

(1) A case study will be done on the benchmark database of Message Spams. This experiment is to demonstrate the use of developed approach for self-organisation of a linguistic attributes hierarchy, given the data.

(2) A set of databases will be tested with the SOLAHs. The performance are used to validate the developed approach to self-organising LAHs by comparing with the performance obtained with the single LDT.

{\bf Experimental environment:} The experiments are carried out on a laptop with 64-bit Windows 10 and x64-based processor with Intel (R) Core (TM) i5-4210U CPU @1.7GHZ 2.4GHZ, 8GB memory. The LAH training algorithm is implemented in C++.

{\bf Discretisation:} For simplicity, all attributes are expressed with three labels, except binary variables, decision or goal variables, which are expressed with the labels as they have.  All neighbouring fuzzy intervals are overlapping with 50\%.

{\bf Mass assignments of attributes:} In terms of Label Semantics, if $n$ labels are used to describe a continuous variable $x$, then there will be $2n-1$ focal sets that are appropriate to describe $x \in \Omega_x$. For discrete variables, the mass assignment on a focal set that contains only one label is 1, but the mass assignment on a focal set that contains two successive labels is 0.

{\bf Ten-fold cross validation:} 90\% of data is used for training, 10\% is used for test. Therefore, data is partitioned to ten parts equally.

{\bf Performance measure:} The performance is measured with:

(1) \emph{Accuracy:} $\mathcal{A} = \frac{TP+TN}{N}$, where, $TP$ is the number of true positive estimates and $TN$ is the number of true negative estimates, and $N$ is the number of samples in the database.

(2) \emph{Area under ROC curve ($AUR$)}: A ROC curve is used to measure how well the classifier separates the two classes without reference to a goal threshold. The area under the ROC curve has been formalized in \cite{He2014}.

(3) \emph{Rule number $\beta$:} As defined in \cite{He2014}, the rule number of an LAH is the sum of branch numbers, extracted from all LDTs in the LAH. Since a probability threshold was introduced during the process of LDT training, a branch training may stop earlier. Hence, the actual branch number in an LDT may be much less than a full LDT.
%Assume $\rho_i$ is the number of input attributes for $LDT_i$ in an LAH. An LDT will be a $\tau$-branch tree with $\tau^{\rho_i}$ branches. The number of total branches in a hierarchy:
%%%%%%%%%%%%%%%%%%%%%%%%%%%%%%%%%%%%%%%%%%%%%%%%%%%%%%%%%%%%%%%%%%%%%
%\begin{align}\label{eq:branches}
%\beta(H)= \sum_i(\beta_i) =\sum_i(\tau^{\rho_i}).
%\end{align}
%%%%%%%%%%%%%%%%%%%%%%%%%%%%%%%%%%%%%%%%%%%%%%%%%%%%%%%%%%%%%%%%%%%%%
%In the implementation, the learning threshold is set to the maximum 1. Namely, if the probability of positive or negative gets 1, the node will be a leave (i.e. no further learning process). Therefore, real number of branches of %an LDT is less than the maximum, and the real number of branches of an LAH is less than the number calculated with Eq.\ref{eq:branches}.

(4) \emph{The running time $t$:} it is the time, spending on the process of ten-fold crossing validation, including training, testing, and overhead for data splitting and exchanging.

\subsection{A case study on a benchmark database}
To demonstrate the developed approach, the first experiment is carried out on the benchmark database, 'SMSSpamCollection'. The performance for the LAH that was constructed by the proposed approach, the LAH in \cite{He2017}, obtained manually,  and the single LDT, is evaluated and compared.

\subsubsection{The database}
The SMSSpamCollection database \cite{Almeida2011}, has 5574 raw messages, including 747 spams. The two sets of features, extracted for the research \cite{He2016,He2017}, are used for the experiments. Table \ref{tab:SMS-20} shows the set of 20 features, and Table \ref{tab:SMS-14} presents the set of 14 features.
\begin{table}[ht!]
\centering
%\begin{center}
\caption{\footnotesize The set of 20 features \cite{He2016}} \label{tab:SMS-20}
\footnotesize
\begin{tabular}{p{12pt}|p{48pt}|p{12pt}|p{88pt}}
\hline
$x$ & key word &   $x$ & key-word\\
\hline\hline
0 & urgent        &  10 &  stop\\
1 & congrat       &  11 &  click\\
2 & !             &  12 &  Text,Txt\\
3 & WIN/WON       &  13 &  sex  \\
4 & Offer         &  14 &  girl \\
5 & Award         &  15 &  cash \\
6 & Prize         &  16 &  free \\
7 & Call          &  17 &  0p, 1p, ..., 9p \\
8 & Reply         &  18 &  EURO, GBP, pound,$\mathcal{L}$, \$, \euro\\
9 & Send          &  19 &  price \\
\hline
\end{tabular}
%\end{center}
\end{table}

%In the 20 attributes of the data set, some attributes may have similar properties. These attributes could be combined together to be one feature, which can be implemented easily with the excel formula: %`if(or($x_1$,$x_2$,...$x_k$),1,0)'. For example, `Award' and `Prize' are combined to be one feature, `award/prize', similarly, `sex/girl', `reply/send', `cash/price',`free/0p..9p'`,`stop/click'. Thus, the number of features is %reduced to 14 from original 20. Hence, a %14-attribute data set is produced.
\begin{table}[htp]
\centering
%\begin{center}
\caption{\footnotesize The set of 14 features \cite{He2017}} \label{tab:SMS-14}
\footnotesize
\begin{tabular}{p{12pt}|p{48pt}|p{12pt}|p{88pt}}
\hline
$x$ & key word &   $x$ & key-word\\
\hline\hline
0 & urgent        &  7 &  call\\
1 & congrat       &  8 &  Reply/send\\
2 & !             &  9 &  stop/click\\
3 & WIN/WON       &  10 & text/txt  \\
4 & Offer         &  11 & cash/price \\
5 & Award/Prize   &  12 & free/0p...9p\\
6 & sex/girl      &  13 &  EURO, GBP, pound,$\mathcal{L}$, \$, \euro\\
\hline
\end{tabular}
%\end{center}
\end{table}
%%%%%%%%%%%%%%%%%%%%%%%%%%%%%%%%%%%%%%%%%%%%%%%%%%%%%%%%%%%%%%%%%%%%%%%%%%%%%%%%%%%%%%%
%%%%%%%%%%%%%%%%%%%%%%%%%%%%%%%%%%%%%%%%%%%%%%%%%%%%%%%%%%%%%%%%%%%%%%%%%%%%%%%%%%%%%%%

\subsubsection{Experimental results on SMS-20}

The six clusters obtained by the proposed distance correlation -based clustering algorithm are:
$S_1$:\{$x_7$,$x_1$,$x_4$,$x_{18}$\}, $S_2$:\{$x_0$,$x_{19}$\}, $S_3$:\{$x_{15}$,$x_{16}$\}, $S_4$:\{$x_6$,$x_2$,$x_3$,$x_8$\}, $S_5$:\{$x_{9}$,$x_{11}$\}, $S_6$:\{$x_5$,$x_{10}$,$x_{12}$,$x_{13}$,$x_{14}$,$x_{17}$\}.

\begin{figure}[htp]
\begin{center}
\includegraphics[width=3.5in]{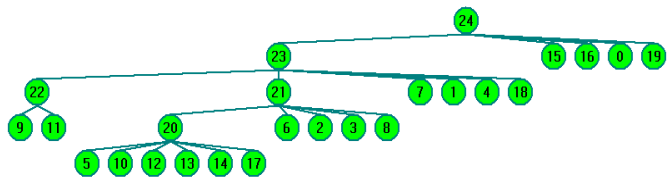}
\caption{The SOLAH on SMS-20} \label{fig:LAH-SMS20}
\end{center}
\end{figure}

\begin{figure}[htp]
\begin{center}
\includegraphics[width=3.5in]{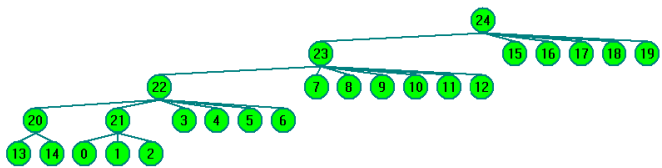}
\caption{A LAH\_m on SMS-20\cite{He2017}}\label{fig:HS-2}
\end{center}
\end{figure}

$S_6$ has the lowest average distance correlation to the decision variable. Hence, the LDT fed by $S_6$ should be at the bottom of the SOLAH, and produces an intermediate attribute $z_{20}$. Here $z$ is used to represent an intermediate attribute. $z_{20}$ with set $S_4$ feeds an LDT in the second level according to the average distance correlation of $S_4$ to decision variable. Sets $S_5$ and $S_4$ have similar average distance correlations to decisional variable. Hence, they sit at the same level in the SOLAH, and their outputs $z_{21}$ and $z_{22}$ feed the third level of LDT constructed with set $S_1$.  Sets $S_2$ and $S_3$ have similar average distance correlations to the decision variable. Hence they are at the same level. As they are at the top level, $S_2$, $S_3$ and the output $z_{23}$ of the LDT at the third level  will be an input of the top LDT. Figs. \ref{fig:LAH-SMS20} and \ref{fig:HS-2}  illustrate the SOLAH and the best $LAH_m$, manually produced in \cite{He2017}. The two LAHs have the same levels and the same number of LDTs. But the compositions are different.
%%%%%%%%%%%%%%%%%%%%%%%%%%%%%%%%%%%%%%%%%%%%%%%%%%%%%%%%%%%%%%%%%%%
%%%%%%%%%%%%%%%%%%%%%%%%%%%%%%%%%%%%%%%%%%%%%%%%%%%%%%%%%%%%%%%%%%%
%%%%%%%%%%%%%%%%%%%%%%%%%%%%%%%%%%%%%%%%%%%%%%%%%%%%%%%%%%%%%%%%%%%
%%%%%%%%%%%%%%%%%%%%%%%%%%%%%%%%%%%%%%%%%%%%%%%%%%%%%%%%%%%%%%%%%%%
%%%%%%%%%%%%%%%%%%%%%%%%%%%%%%%%%%%%%%%%%%%%%%%%%%%%%%%%%%%%%%%%%%%
%%%%%%%%%%%%%%%%%%%%%%%%%%%%%%%%%%%%%%%%%%%%%%%%%%%%%%%%%%%%%%%%%%%
Table \ref{tab:LAH-SMS20-perf} shows the performance for the SOLAH, the $LAH_m$ in \cite{He2017} and the single LDT fed by the whole feature vector (denoted as $LDT(\vec{x})$. It can be seen that the accuracy $A$ of $SOLAH$ is very close to the performance of $LAH_m$ and $LDT(\vec{x})$, and the performance of the AUC is slightly smaller than that of $LAH_m$, but better than that of $LDT(\vec{x})$.  However, $SOLAH$ has the smallest branch number $beta$ and running time $T$, which is the time of ten-fold crossing validation on a solution, measured in milliseconds (ms).
%%%%%%%%%%%%%%%%%%%%%%%%%%%%%%%%%%%%%%%%%%%%%%%%%%%%%%%%%%%%%%%%%%%
%%%%%%%%%%%%%%%%%%%%%%%%%%%%%%%%%%%%%%%%%%%%%%%%%%%%%%%%%%%%%%%%%%%
\begin{table}[htp]
\centering
\caption{\footnotesize The performance of solutions} \label{tab:LAH-SMS20-perf}
\footnotesize
\begin{tabular}{l|p{28pt}|p{28pt}|p{28pt}|p{28pt}}
\hline
LAHs & A &$AUC$&$\beta$ & $T$(ms)\\
\hline\hline
$SOLAH$, Fig. \ref{fig:LAH-SMS20}         & 0.936491 & 0.946402&197&101766\\
$LAH_m$\cite{He2017}, Fig. \ref{fig:HS-2} & 0.948511 & 0.949233 &267& 143718 \\
$LDT(\bar{x})$                            & 0.956225 & 0.895815&775&1339234 \\
\hline
\end{tabular}
\end{table}
%Fig. \ref{fig:ROC-SMS20} shows the ROC curves of the solutions in Table \ref{tab:performace-SMS20}.
%It can be seen that the self-organised and manually conducted LAHs have better ROC performance than the single LDT.
%%%%%%%%%%%%%%%%%%%%%%%%%%%%%%%%%%%%%%%%%%%%%%%%%%%%%%%%%%%%%%%%%%%
%%%%%%%%%%%%%%%%%%%%%%%%%%%%%%%%%%%%%%%%%%%%%%%%%%%%%%%%%%%%%%%%%%%
\subsubsection{Experimental results on SMS-14}

For this experiment, the cluster number is preset to 4. But the clusters obtained by the DCC algorithm are: $S_1$:\{$x_6$,$x_11$\}, $S_2$:\{$x_9$,$x_{4}$,$x_8$,$x_{10}$\}, $S_3$:\{$x_{5}$,$x_{0}$,$x_{1}$,$x_{2}$,$x_{3}$,$x_{12}$,$x_{13}$\}.
Set $S_1$ has the lowest average distance correlation to the decision variable, thus the LDT fed by $S_1$ is placed  at the bottom of the SOLAH. The LDT fed by set $S_1$ and the output of the first level at the second level of the SOLAH, and the LDT fed by set $S_3$ sits at the top level of the LAH.
%LAH-SMS14-1
%14	6	11	
%15	2	1	4	7	8	
%16	15	14	5	0	3	13	
%17	16	9	10	12	

%LAH-SMS14-2
%[6;11]
%[5;0;3;7;13]
%[9;10;12]
%[4;1;2;8]

\begin{figure}[htp]
\begin{center}
\includegraphics[width=3.2in]{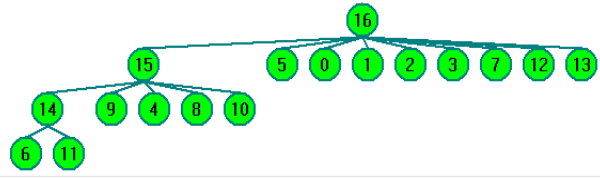}
\caption{The SOLAH on SMS-14} \label{fig:LAH-SMS14}
\end{center}
\end{figure}

\begin{figure}[htp]
\begin{center}
\includegraphics[width=3.2in]{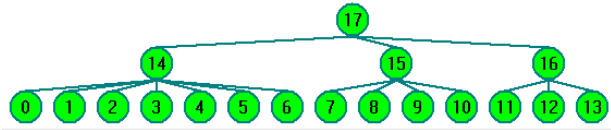}
\caption{A LAH on SMS-14, obtained manually in \cite{He2017}}\label{fig:HH3}
\end{center}
\end{figure}

Figs. \ref{fig:LAH-SMS14} and \ref{fig:HH3}  illustrate the SOLAH and the $LAH_m$ with the best $AUC$, which was obtained by the attribute composition 3 in \cite{He2017}. The two LAHs have the same levels, and the attributes are decomposed to 3 clusters. But the compositions of LAHs are different. In the $LAH_m$, all LDTs fed by the clusters of input attributes are placed at the bottom of the LAH. Hence a new LDT, fed by all intermediate attributes, which were produced by the bottom level of LDTs, is added on the top of the LAH. $SOLAH$ has higher level than $LAH_m$, and all LDTs fed by the attribute clusters are cascaded.

Table \ref{tab:LAH-SMS14} shows the performance for the SOLAH, the $LAH_m$, constructed with composition 3 manually in \cite{He2017} and the single LDT ($LDT(\vec{x})$). All solutions in Table \ref{tab:LAH-SMS14} obtained similar performance in accuracy and the area under ROC. All performance values of $SOLAH$ are in between $LAH_m$ and $LDT(\bar{x})$.

\begin{table}[htp]
\centering
\caption{\footnotesize The performance of solutions} \label{tab:LAH-SMS14}
\footnotesize
\begin{tabular}{l|p{28pt}|p{28pt}|p{28pt}|p{28pt}}
\hline
LAHs & A &$AUC$&$\beta$ & $T$(ms)\\
\hline\hline
$SOLAH$, Fig. \ref{fig:LAH-SMS20}      & 0.955328 &   0.943391&339& 101766\\
$LAH_m$\cite{He2017}, Fig. \ref{fig:HS-2} & 0.951202 &   0.957466  &130& 79719 \\
$LDT(\bar{x})$                   & 0.961607  &  0.924536&583&613704 \\
\hline
\end{tabular}
\end{table}

\subsubsection{Impact of the preset cluster number}

Similar to classic $k$-means clustering algorithm, the preset cluster number $k$ is important for the construction of LAHs, and thus has important impact on the performance of decision making or classification. Here, the relationships between $k$ and structures of LAHs and between $k$ and performance are observed. It should be noticed that the real cluster number may be different to the preset cluster number, as  the present cluster number decides the range ($\alpha$) of distance correlation of a cluster $C$, which is formed with the attributes that have the distance correlations in the range $\alpha$ to an attribute $x$, which has the largest distance correlation in the distance correlation matrix.  The rest attributes could have larger distance correlations than the members except attribute $x$ in cluster $C$.
Table \ref{tab:LAH-SMS14-1} shows the solutions for $k$= 2...10. It can be seen that the real cluster numbers are different to the preset cluster number $k$ in some cases. The cluster numbers varies from 3 to 6. All SOLAHs have similar performance. The SOLAH constructed with 3 clusters obtained the highest accuracy, but it has more branches. The SOLAHs constructed with 4 clusters have the highest $AUC$ values. The clusters obtained when $k=7$ is the same as that when $k$=8, which is happened when $k=9,10$ as well. In Table \ref{tab:LAH-SMS14-1}, $\beta$ denotes the branch number, $\iota$ denotes the level of the hierarchy.
\begin{table}[htp]
\centering
\caption{\footnotesize The properties of LAHs on SMS14} \label{tab:LAH-SMS14-1}
\footnotesize
\begin{tabular}{p{10pt}|p{10pt}|p{28pt}|p{28pt}|p{28pt}|p{10pt}|p{28pt}}
\hline
$k$  &$K$   & A     &$AUC$  &$\beta$& $\iota$&$T$(ms)\\
\hline\hline
2    & 3    & 0.956 & 0.946 &  339  & 4 & 221547\\
3    & 4    & 0.937 & 0.953 &  184  & 4 & 101875\\
4    & 4    & 0.936 & 0.950 &  154  & 4 & 84093\\
5    & 6    & 0.932 & 0.949 &  159  & 5 & 91375\\
6    & 6    & 0.932 & 0.948 &  154  & 6 & 89828\\
7,8  & 6    & 0.932 & 0.945 &  108  & 6 & 63313\\
9,10 & 6    & 0.932 & 0.941 &  130  & 5 & 76859\\
\hline
\end{tabular}
\end{table}

Similarly, Table \ref{tab:LAH-SMS20-K} lists the performance of SOLAHs constructed with various $k$ values from 2 to 10. The AUC values of all SOLAHs are very close. When the preset $k=2$, the SOLAH was constructed with two clusters. It obtained the highest accuracy, but lowest AUC value and the largest branch number. When the preset $k$=9,10, the same cluster sets were produced, with which, the SOLAH was produced.  It has the lowest accuracy, but the smallest branch number. For all preset $k$=3...6, different sets of 5 clusters were produced, with which, four different SOLAHs were produced. They have the same level, and their performance values ($A$ and AUC) are very close. When the preset $k$=5,6,7, the produced SOLAHs have the same performance values ($A$ and AUC). But they have different branch numbers.
\begin{table}[htp]
\centering
\caption{\footnotesize The properties of LAHs on SMS20} \label{tab:LAH-SMS20-K}
\footnotesize
\begin{tabular}{p{10pt}|p{10pt}|p{28pt}|p{28pt}|p{28pt}|p{10pt}|p{28pt}}
\hline
$k$  &$K$   & A     &$AUC$  &$\beta$& $\iota$&$T$(ms)\\
\hline\hline
2   & 2 &  0.957  &  0.933   &540   & 3 & 467328\\
3   & 5 &  0.934  &  0.950   &189   & 5 & 111016\\
4   & 5 &  0.949  &  0.941   &203   & 5 & 114265\\
5   & 5 &  0.936  &  0.946   &187   & 5 & 101594\\
6   & 5 &  0.936  &  0.946   &197   & 5 & 107766\\
7   & 6 &  0.936  &  0.946   &204   & 5 & 115750\\
8   & 7 &  0.929  &  0.940   &163   & 6 & 93891\\
9,10& 7 &  0.891  &  0.944   &157   & 7 & 104938\\
\hline
\end{tabular}
\end{table}

From the data on both SMS20 and SMS14, it can be seen that a suitable cluster size $K$ will have a good trade-off on accuracy performance and time complexity (branch number).
%%%%%%%%%%%%%%%%%%%%%%%%%%%%%%%%%%%%%%%%%%%%%%%%%%%%%%%%%%%%%%%%%%%%%
%%%%%%%%%%%%%%%%%%%%%%%%%%%%%%%%%%%%%%%%%%%%%%%%%%%%%%%%%%%%%%%%%%%%%
\subsection{Validation on some benchmark databases}
\subsubsection{The 12 benchmark databases}
The experiments on 12 benchmark databases from UCI machine learning repository are conducted for validating the proposed approach of constructing LAHs. Table \ref{tab:datasets} provides the basic properties of the 12 data sets, including, database name, attribute number ($n$), goal state number ($N_g$), total sample number ($N$), class distribution ($N_c$).
%%%%%%%%%%%%%%%%%%%%%%%%%%%%%%%%%%%%%%%%%%%%%%%%%%%%%%%%%%%%%%%%%%%
%%%%%%%%%%%%%%%%%%%%%%%%%%%%%%%%%%%%%%%%%%%%%%%%%%%%%%%%%%%%%%%%%%%
\begin{table}[ht!]
\begin{center}
\caption{\footnotesize Database properties} \label{tab:datasets}
\footnotesize
\begin{tabular}{l|l|l|l|l}
\hline
Datasets                  &$n$  & $N_g$   & $N$   & $N_c$\\
\hline\hline
Beast Cancer (BC)		  & 9   &    2    &  286  & 201,85\\
Wisconsin BC              & 9   &    2    &  569  & 357,212\\
Ecoli		              & 7   &    8    &  336  & 143,77,52,35,20,5,2,2\\
Glass		              & 9   &    6    &  214  & 70,17,76,13,9,29\\
Heart-C	                  & 13  &    2    &  303  & 165,138\\
Heart-statlog             & 13  &    2    &  270  & 150,120\\
Hepatitis	              & 19  &    2    &  155  & 70,85\\
Wine		              & 13  &    3    &  178  & 59,71,48\\
Liver Disorders           & 6   &    2    &  346  & 146,200\\
Diabetes                  & 8   &    2    &  768  & 500,268\\
Ionoshpere	              & 33  &    2    &  351  & 231,120\\
Sonar		              & 60  &    2    &  208  & 97,111\\
\hline
\end{tabular}
\end{center}
\end{table}
%%%%%%%%%%%%%%%%%%%%%%%%%%%%%%%%%%%%%%%%%%%%%%%%%%%%%%%%%%%%%%%%%%%
%%%%%%%%%%%%%%%%%%%%%%%%%%%%%%%%%%%%%%%%%%%%%%%%%%%%%%%%%%%%%%%%%%%		

\subsubsection{Comparisons with LDT}
%%%%%%%%%%%%%%%%%%%%%%%%%%%%%%%%%%%%%%%%%%%%%%%%%%%%%%%%%%%%%%%%%%%
%%%%%%%%%%%%%%%%%%%%%%%%%%%%%%%%%%%%%%%%%%%%%%%%%%%%%%%%%%%%%%%%%%%
%%%%%%%%%%%%%%%%%%%%%%%%%%%%%%%%%%%%%%%%%%%%%%%%%%%%%%%%%%%%%%%%%%%
%%%%%%%%%%%%%%%%%%%%%%%%%%%%%%%%%%%%%%%%%%%%%%%%%%%%%%%%%%%%%%%%%%%
Table \ref{tab:comparisonLDT} shows the performance of SOLAHs and corresponding single LDTs on the 12 benchmark data sets. It can be seen that the SOLAH achieved dominated better performance in accuracy $A$ and $AUC$ than the single LDT for most databases. For databases, Ecoli and Liver, SOLAHs achieve similar accuracy to LDTs, but their performance in $AUC$ is better than that of LDTs. The SOLAHs for most databases have less branches than the single LDT trained by the same database.

%%%%%%%%%%%%%%%%%%%%%%%%%%%%%%%%%%%%%%%%%%%%%%%%%%%%%%%%%%%%%%%%%%%
%%%%%%%%%%%%%%%%%%%%%%%%%%%%%%%%%%%%%%%%%%%%%%%%%%%%%%%%%%%%%%%%%%%
\begin{table}[ht!]
\begin{center}
\caption{\footnotesize Performance comparison between constructed LAHs
and the single LDTs for different databases, the unit of time (T) is second} \label{tab:comparisonLDT}
\footnotesize
\begin{tabular}{p{30pt}|p{10pt}|p{12pt}|p{14pt}|p{5pt}|p{5pt}|p{5pt}|p{10pt}|p{12pt}|p{14pt}|p{10pt}}
\hline &\multicolumn{6}{|l|}{SOLAH}  &\multicolumn{4}{|l}{LDT}\\
\cline{2-11}
\raisebox{1.5ex}{Databases}&$A$&\small{AUC}&$\beta$&$T$&$\iota$&$K$& $A$&\small{AUC}&$\beta$&$T$\\
\hline \hline
BreastC.      &0.78    &0.84  &985        &9    &4    &3      &0.70  &0.64  &3201    &66  \\
WBC           &0.97    &0.98  &1100       &36  &3    &2      &0.95  &0.96  &1029     &45  \\
Ecoli         &0.85    &0.93  &1103       &15  &4    &3      &0.85  &0.89  &3337     &12  \\
Glass         &0.81    &0.91  &5124       &48  &3    &2      &0.68  &0.83  &9433     &156 \\
Heart-c       &0.81    &0.81  &1452       &24  &5    &4      &0.76  &0.78  &1345     &40  \\
Heart-s.      &0.79    &0.83  &972        &11  &5    &4      &0.77  &0.80  &1053     &24  \\
Hepatitis     &0.92    &0.94  &4278       &47  &4    &6      &0.79  &0.82  &1677     &45   \\
Wine          &1.00    &0.99  &929        &6   &5    &5      &0.94  &0.98  &1309     &26  \\
Liver         &0.56    &0.60  &226        &3   &3    &2      &0.57  &0.56  &1585     &24  \\
Diabetes      &0.76    &0.82  &1018       &33  &3    &2      &0.74  &0.79  &14865    &667 \\
Ionosphere    &0.87    &0.71  &4393       &14  &5    &9      &0.87  &0.87  &12141    &6953 \\
Sonar         &0.73    &0.80  &26328      &52  &6    &22     &0.67  &0.73  &9253     &4968 \\
\hline
\end{tabular}
\end{center}
\end{table}

As databases Ionosphere and Sonar have 33 and 60 attributes, respectively, the ten-fold crossing validation of a single LDT on the two databases is not acceptable. Therefore, the two-fold crossing validation of single LDTs on the two high dimension databases are performed. For comparison, the two-fold crossing validation of the constructed LAHs on the two databases are carried out as well. The results are provided in Table \ref{tab:comparison}. For database Ionosphere, the LAH has the same accuracy as the LDT, but the area under ROC obtained by the SOLAH is smaller than that by the single LDT. Namely, the half size of data may not be enough to train the SOLAH. However, the running times of SOLAHs for the two databases are 14s and 52s respectively, which are much less than that of LDTs for the two databases. The database Sonar provides a very interesting case: the SOLAH has 26328 branches, while the single LDT has only 9253 branches. Regarding the branch number, the complexity of the SOLAH model is worse than the single LDT. But the running time for SOLAH is 52s, while 4968s for LDT. This is because that the computing time is not only related to branch number, but also related to branch length. The computing complexity can be $O(\beta\times l)$. It can be seen that the SOLAH for the Sonar database has 22 layers, which means that at least 22 LDTs are embedded in the SOLAH. There are total 60 input attributes, plus 22 intermediate attributes, the average input attributes of an LDT in the SOLAH is 3.68. Hence, the average length of branch is 3,68. We can roughly say that the running time of the SOLAH is proportional to $26328\times 3.68$ = 96887, but the single LDT has 4968 branches, the length of which is 60, then the running time of the single LDT is proportional to $4968\times 60$ = 298080, which is much larger than the figure for the SOLAH.

In order to observe the performance improvement of the LAH that is trained by 90\% of data when ten-fold crossing validation is applied, Table \ref{tab:highDims} particularly lists the performance of the SOLAHs with ten-fold crossing validation on the two high dimensional databases. The performance in both accuracy and the area under ROC is improved, especially for the database, Ionosphere, the performance in accuracy and the area under ROC are improved very much, compared to the SOLAH with two-fold crossing validation. Moreover, the running times of the SOLAHs for the two databases are 143s and 507s, respectively,  which are much shorter than that of the single LDT with two-fold crossing validation in Table \ref{tab:comparisonLDT}.

\begin{table}[ht!]
\begin{center}
\caption{\footnotesize Performance of SOLAHs on the two high dimensional databases} \label{tab:highDims}
\footnotesize
\begin{tabular}{l|l|l|l|l|l|l}
\hline
Databases&$A$&$AUC$&$\beta$&$T(ms)$&$\iota$&$K$\\
\hline \hline
Ionosphere    &0.95    &0.98  &7717       &143  &5    &9\\
Sonar         &0.78    &0.84  &33820      &507  &6    &22\\
\hline
\end{tabular}
\end{center}
\end{table}

%%%%%%%%%%%%%%%%%%%%%%%%%%%%%%%%%%%%%%%%%%%%%%%%%%%%%%%%%%%%%%%%%%%
%%%%%%%%%%%%%%%%%%%%%%%%%%%%%%%%%%%%%%%%%%%%%%%%%%%%%%%%%%%%%%%%%%%
%%%%%%%%%%%%%%%%%%%%%%%%%%%%%%%%%%%%%%%%%%%%%%%%%%%%%%%%%%%%%%%%%%%
%\vspace{-18pt}
\subsubsection{Comparisons with other machine learning algorithms}
The average accuracy and standard deviations of SOLAH for ten runs on the 12 data sets
from UCI machine learning repository are further compared with the optimal LAH, obtained by a Genetic Algorithm Wapper in \cite{He2014}, as well as three well-known machine learning algorithms, such as C4.5, Naive Bayes (NB) and Neural Networks (NN). The performance of these algorithms WAS evaluated with WEKA \cite{WF00} by Qin and Lawry in
\cite{Qin2005a}, where, WEKA \cite{WF00} was used to generate the results of J48 (C4.5 in WEKA) unpruned tree, Naive Bayes and Neural Networks with default parameter settings for ten runs of 50\%-50\% splitting training and test data.

Table \ref{tab:comparison} shows the performance of SOLAHs and the existing results in literature for the 12 data sets. SOLAH wins 4 data sets, which is next to OptLAH, which wins 5 data sets.  SOLAH obtains comparable performance in accuracy, compared to other algorithms, and even achieves better performance than the LAHs, optimised by GAW, for some tested data sets.
%%%%%%%%%%%%%%%%%%%%%%%%%%%%%%%%%%%%%%%%%%%%%%%%%%%%%%%%%%%%%%%%%%%
%%%%%%%%%%%%%%%%%%%%%%%%%%%%%%%%%%%%%%%%%%%%%%%%%%%%%%%%%%%%%%%%%%%
\begin{table}[ht!]
\begin{center}
\caption{\footnotesize Average accuracies (\%) and standard deviations
obtained by C4.5, NB, NN, OptLAH and SOLAHs}\label{tab:comparison}
\footnotesize
\begin{tabular}{p{36pt}|p{45pt}|p{45pt}|p{45pt}|p{45pt}|p{45pt}}
\hline
$Databases$ &C4.5 &N.B.&N.N.& OptLAH & SOLAH\\
\hline \hline
%Balance &3 & 4 & 625&No. &$79.20\pm1.53$ &$89.46\pm2.09$ &$90.38\pm1.18$  &$79.84\pm0.00$ \\
BreastC.      &$69.16\pm4.14$ &$71.26\pm2.96$ &$66.50\pm3.48$  &$71.77\pm2.06$&${\bf75.17}\pm4.30 $ \\
WBC           &$94.38\pm1.42$ &$96.28\pm0.73$ &$94.96\pm0.80$  &${\bf96.67}\pm0.20$&$94.85\pm0.25 $ \\
Ecoli         &$78.99\pm2.23$ &$85.36\pm2.42$ &$82.62\pm3.18$  &$84.02\pm0.92$&${\bf86.11}\pm0.91 $ \\
Glass         &$64.77\pm5.10$ &$45.99\pm7.00$ &$64.30\pm3.38$  &${\bf71.31}\pm1.75$&$71.26\pm8.08 $ \\
Heart-c       &$75.50\pm3.79$ &${\bf84.24}\pm2.09$ &$79.93\pm3.99$  &$82.81\pm4.25$&$78.38\pm3.50 $ \\
Heart-s.      &$75.78\pm3.16$ &$84.00\pm1.68$ &$78.89\pm3.05$  &${\bf84.85}\pm2.31$&$78.15\pm1.05 $ \\
Hepatitis     &$76.75\pm4.68$ &$83.25\pm3.99$ &$81.69\pm2.48$  &${\bf94.84}\pm1.01$&$85.15\pm9.12 $ \\
Ionosphere    &$89.60\pm2.13$ &$82.97\pm2.51$ &$87.77\pm2.88$  &$89.80\pm1.63$&${\bf92.12}\pm4.03$ \\
Liver         &$65.23\pm3.86$ &$55.41\pm5.39$ &${\bf66.74}\pm4.89$  &$58.46\pm0.76$&$56.16\pm0.90 $ \\
Diabetics     &$72.16\pm2.80$ &$75.05\pm2.37$ &$74.64\pm1.41$  &${\bf76.07}\pm1.33$&$75.09\pm1.39 $ \\
Sonar         &$70.48\pm0.00$ &$70.19\pm0.00$ &${\bf81.05}\pm0.00$  &$74.81\pm4.81$&$75.72\pm3.74 $ \\
Wine          &$88.09\pm4.14$ &$96.29\pm2.12$ &$96.85\pm1.57$  &$97.58\pm0.27$&${\bf98.17} \pm2.95 $ \\
\hline
\end{tabular}
\end{center}
\end{table}

\section{Conclusions}
In this paper, the contribution of the innovative research are summarised as below:

(1) We proposed heuristic algorithm to construct a linguistic attribute hierarchy for semantic attribute deep learning in decision making or classification. The self-constructed linguistic attribute hierarchy provides a new form of deep learning, in contrast to conventional deep learning.

(2) The proposed algorithm for the self-organisation of an LAH is much more efficient than meta-heuristic algorithm, and the self-organised linguistic attribute hierarchy
can obtain the fusion performance better than or comparable to the single linguistic decision tree, fed with the full set of attributes.

(3) The most important is that the heuristical self-organisation of such linguistic attribute hierarchy can effectively solve the ’curse of dimensionality’ in machine learning, which is critical challenge in the implementation of IoT intelligence. Hence, the research results will promote a wider of applications of the linguistic attribute hierarchy in big data analysis and IoT intelligence.

(4) A linguistic attribute hierarchy, embedded with linguistic decision trees, will provide a transparent hierarchical decision making or classification. Hence, it could help us to look insight of the decision making process for different purposes (e.g. the effect of adversary samples in decision making).

(5) Comparing with other machine learning in literature, the self-organised LAH obtains comparable performance on the tested data sets, and even achieved better performance than the optimal LAHs, obtained by GAW for some tested data sets.

We will implement the LAH on an embedded system (e.g. a Raspberry Pi system) for a specific task, and will further improve the algorithms and develop new algorithms to construct high efficient and effective linguistic attribute hierarchy, embedded with other machine learning models for decision making or classification in future.
%%%%%%%%%%%%%%%%%%%%%%%%%%%%%%%%%%%%%%%%%%%%%%%%%%%%%%%%%%%%%%%%%%%
%%%%%%%%%%%%%%%%%%%%%%%%%%%%%%%%%%%%%%%%%%%%%%%%%%%%%%%%%%%%%%%%%%%
%\bibliographystyle{kapalike}

\section*{Appendix}
\begin{figure}[htp]
\begin{center}
\includegraphics[width=3.0in]{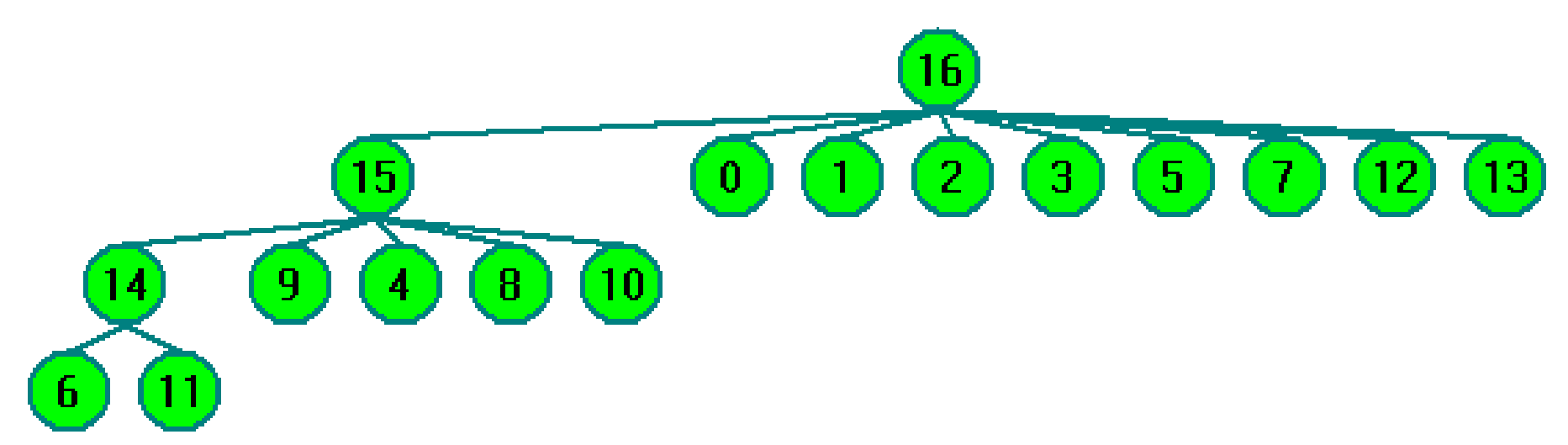}
\caption{LAH based on SMS14 with $k$ = 2} \label{fig:LAH14-2}
\end{center}
\end{figure}

\begin{figure}[htp]
\begin{center}
\includegraphics[width=3.0in]{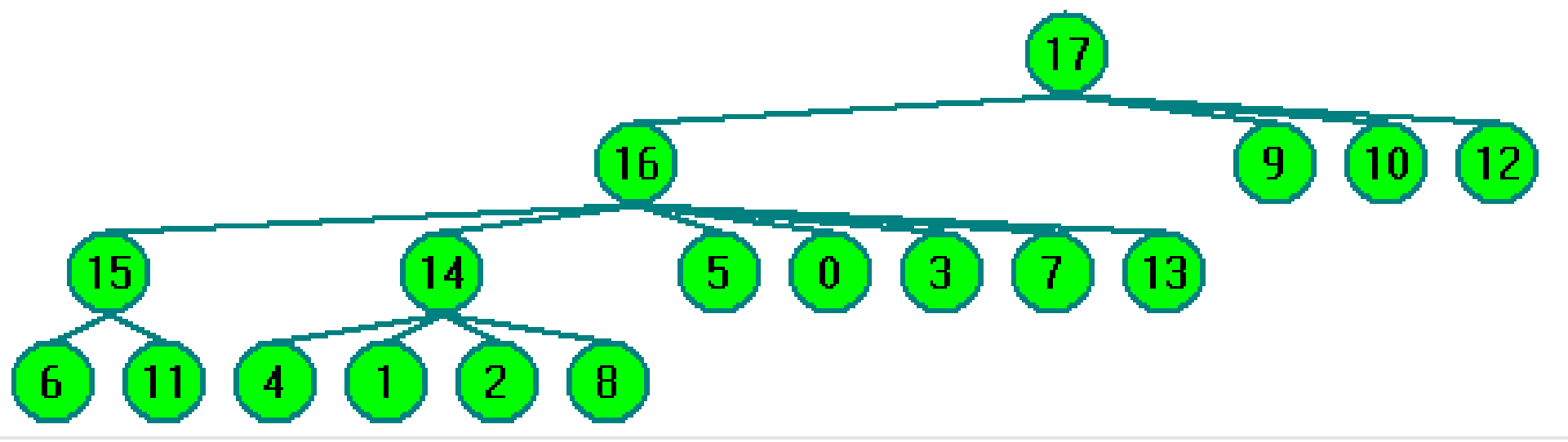}
\caption{LAH based on SMS14 with $k$ = 3} \label{fig:LAH14-3}
\end{center}
\end{figure}

\begin{figure}[htp]
\begin{center}
\includegraphics[width=3.0in]{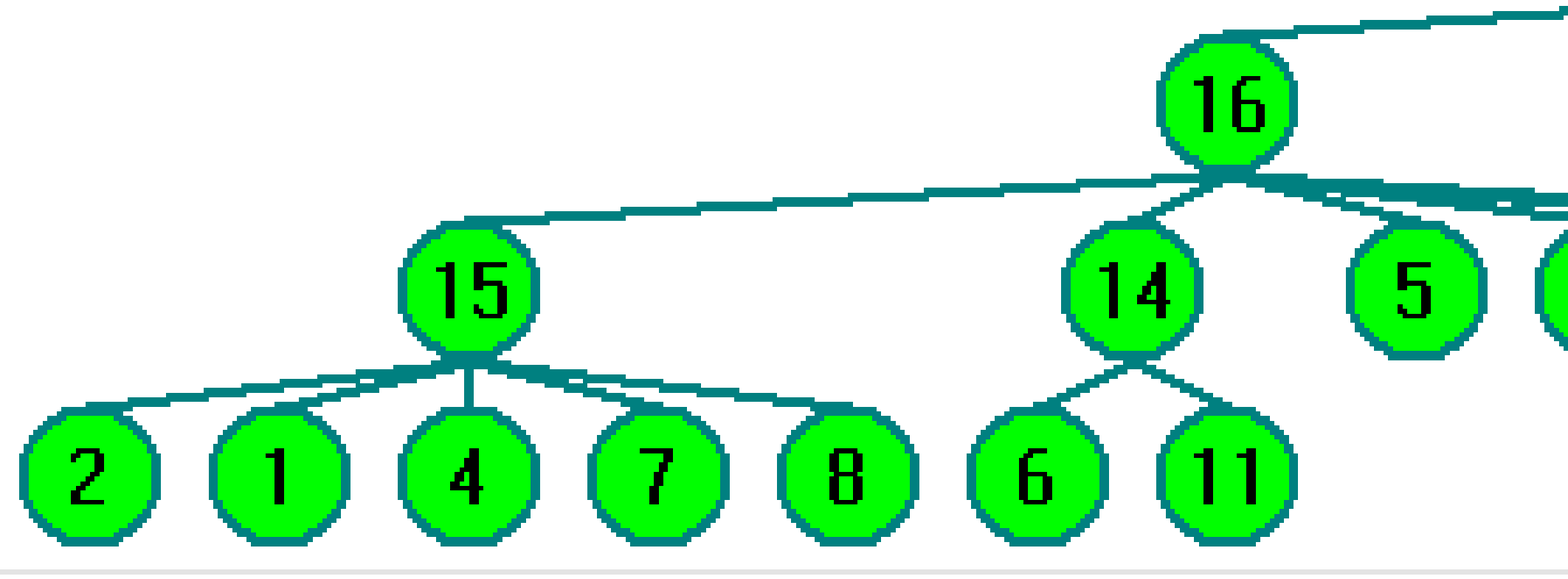}
\caption{LAH based on SMS14 with $k$ = 4} \label{fig:LAH14-4}
\end{center}
\end{figure}

\begin{figure}[htp]
\begin{center}
\includegraphics[width=3.0in]{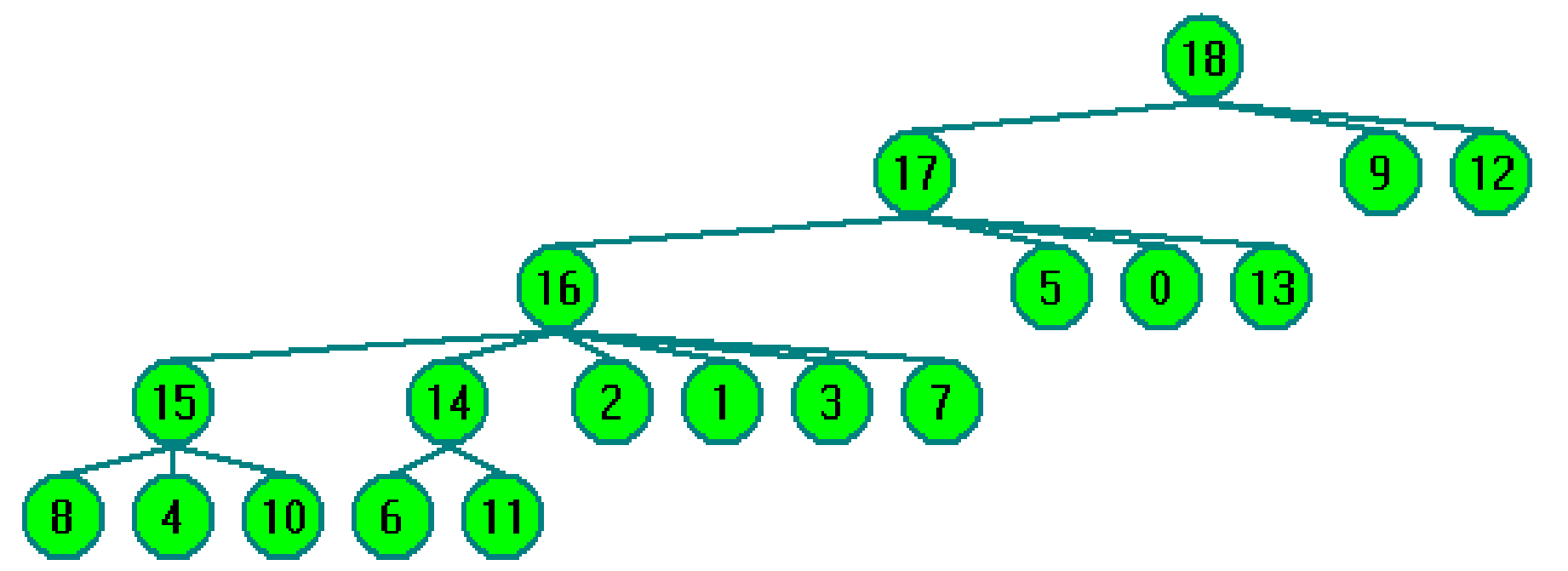}
\caption{LAH based on SMS14 with $k$ = 5} \label{fig:LAH14-5}
\end{center}
\end{figure}

\begin{figure}[htp]
\begin{center}
\includegraphics[width=3.0in]{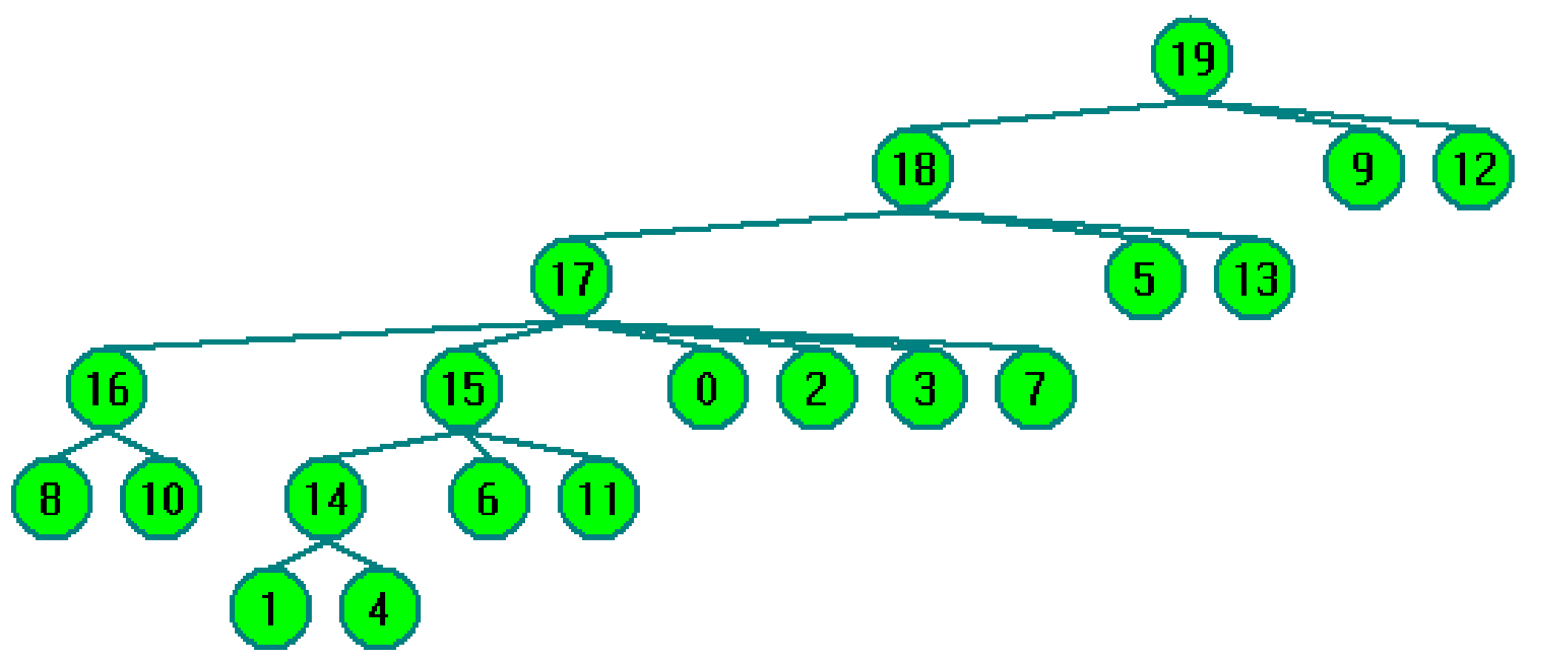}
\caption{LAH based on SMS14 with $k$ = 6} \label{fig:LAH14-6}
\end{center}
\end{figure}

\begin{figure}[htp]
\begin{center}
\includegraphics[width=3.0in]{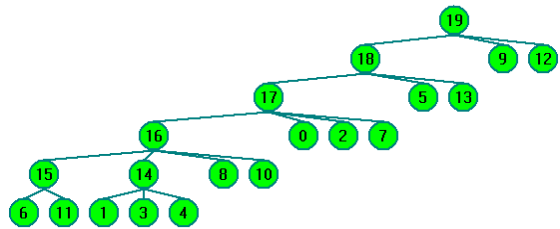}
\caption{LAH based on SMS14 with $k$ = 7 and $k$ = 8.} \label{fig:LAH14-7}
\end{center}
\end{figure}

\begin{figure}[htp]
\begin{center}
\includegraphics[width=3.0in]{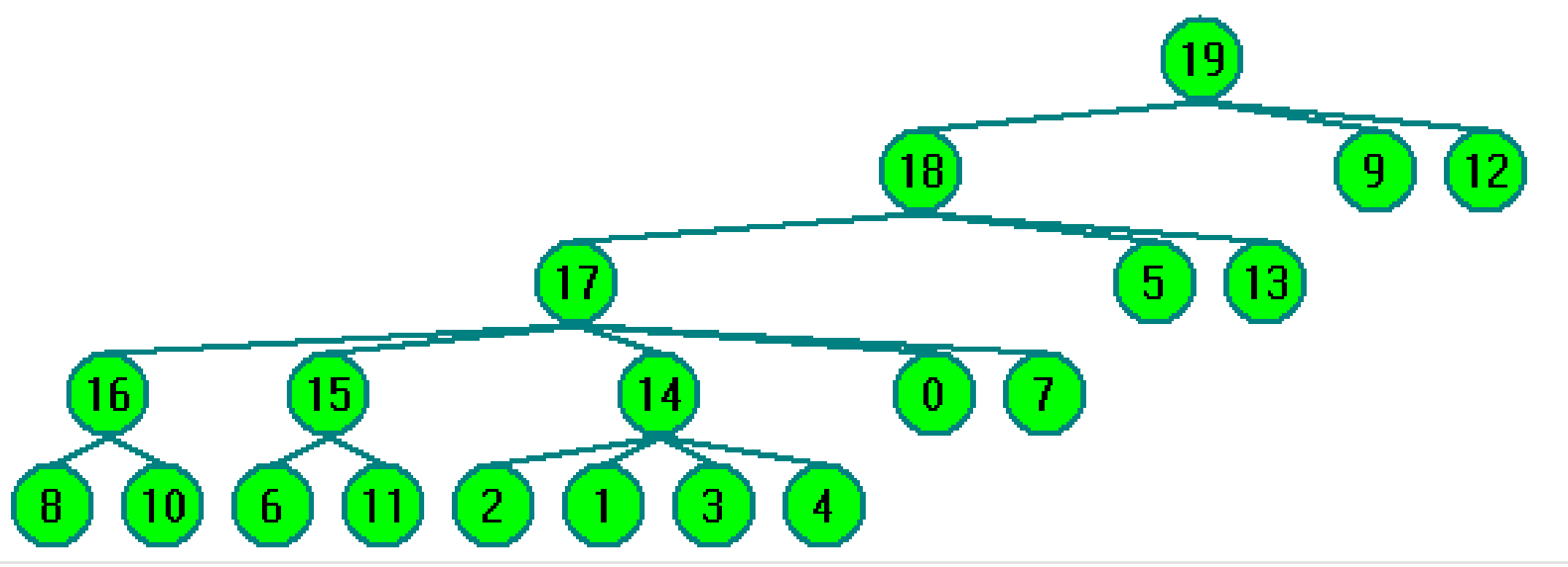}
\caption{LAH based on SMS14 with $k$ = 9 and $k$=10.} \label{fig:LAH14-9}
\end{center}
\end{figure}

\subsection{14 attribute-based LAHs for k = 2..10 (Figs. \ref{fig:LAH14-2}-\ref{fig:LAH14-9})}

\begin{figure}[htp]
\begin{center}
\includegraphics[width=3.0in]{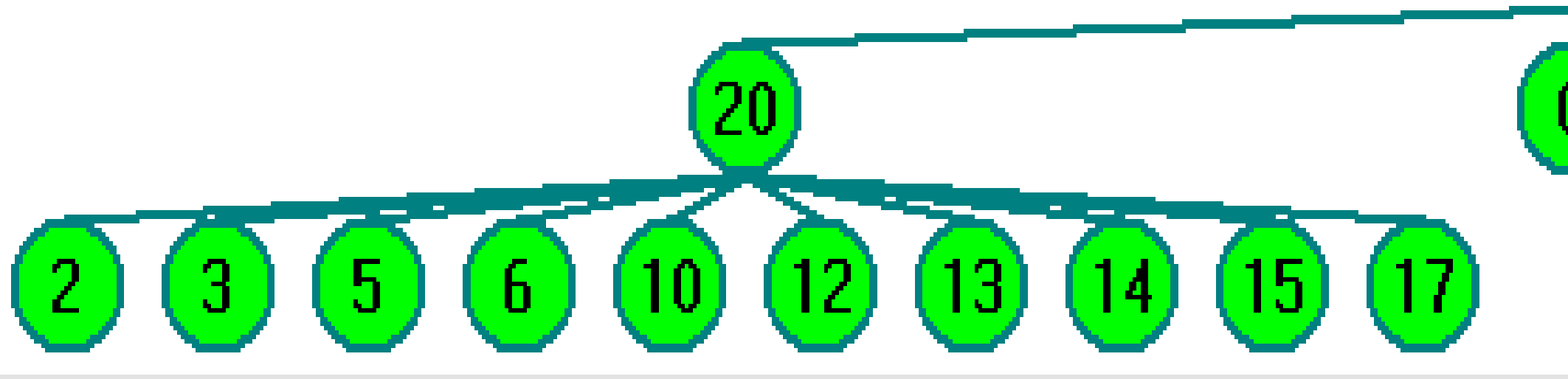}
\caption{LAH for k=2.} \label{fig:LAH20-2}
\end{center}
\end{figure}

\begin{figure}[htp]
\begin{center}
\includegraphics[width=3.0in]{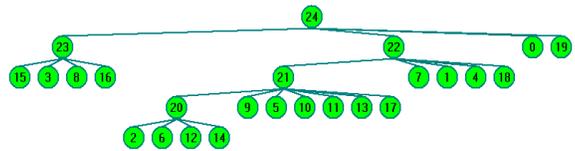}
\caption{LAH for k=3} \label{fig:LAH20-3}
\end{center}
\end{figure}

\begin{figure}[htp]
\begin{center}
\includegraphics[width=3.0in]{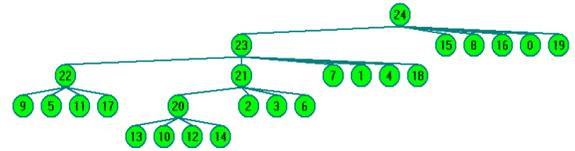}
\caption{LAH for k=4.} \label{fig:LAH20-4}
\end{center}
\end{figure}

\begin{figure}[htp]
\begin{center}
\includegraphics[width=3.0in]{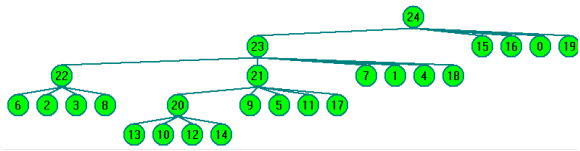}
\caption{LAH for k=5.} \label{fig:LAH20-5}
\end{center}
\end{figure}

\begin{figure}[htp]
\begin{center}
\includegraphics[width=3.0in]{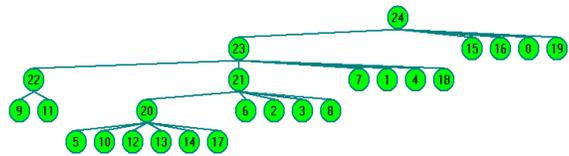}
\caption{LAH for k=6.} \label{fig:LAH20-6}
\end{center}
\end{figure}

\begin{figure}[htp]
\begin{center}
\includegraphics[width=3.0in]{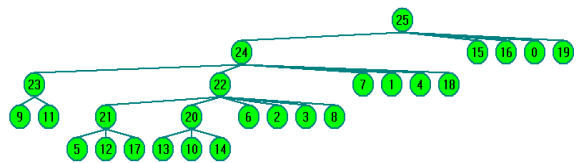}
\caption{LAH for k=7.} \label{fig:LAH20-7}
\end{center}
\end{figure}

\begin{figure}[htp]
\begin{center}
\includegraphics[width=3.0in]{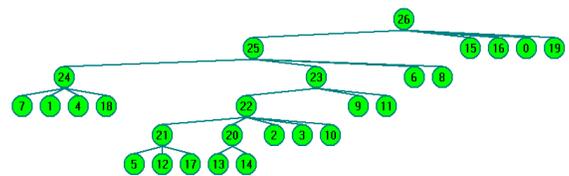}
\caption{LAH for k=8.} \label{fig:LAH20-8}
\end{center}
\end{figure}

\begin{figure}[htp]
\begin{center}
\includegraphics[width=3.0in]{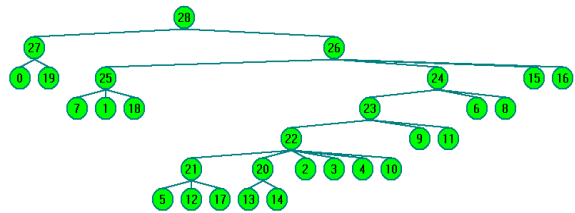}
\caption{LAH for $k$=9 and $k$=10} \label{fig:LAH20-9}
\end{center}
\end{figure}

\subsection{LAHs based on 20 attributes for k=1,..,10 (Figs. \ref{fig:LAH20-2}-\ref{fig:LAH20-9})}

\end{document}